%% file: root.tex
\documentclass[letterpaper, 10 pt, conference]{ieeeconf}  

\IEEEoverridecommandlockouts                              

\input{commands/preamble}
\input{commands/tikz}
\input{commands/acronyms}

\input{commands/math}

\input{commands/notation}

\input{commands/todos}
\usetikzlibrary{external}
\title{\LARGE \bf
Safe and Efficient Path Planning under Uncertainty via\\ Deep Collision Probability Fields
}

\author{Felix Herrmann$^{\dagger,1}$, Sebastian Zach$^{\dagger,1}$, Jacopo Banfi, Jan Peters$^{1,3,4}$, Georgia Chalvatzaki$^{\ddagger,1,3}$ and Davide Tateo$^{\ddagger,1}$
\thanks{$\dagger$ Equal contribution, $\ddagger$ Equal supervision}
\thanks{$^{1}$ Computer Science department, TU Darmstadt}
\thanks{$^{3}$ Hessian.AI}%
\thanks{$^{4}$ German Research Center for AI (DFKI), Research Department: Systems AI for Robot Learning}%
}

\begin{document}

\let\oldtwocolumn\twocolumn
\renewcommand\twocolumn[1][]{%
    \oldtwocolumn[{#1}{
    \begin{center}
    \vspace{-1cm}
    \captionsetup{type=figure}
    \input{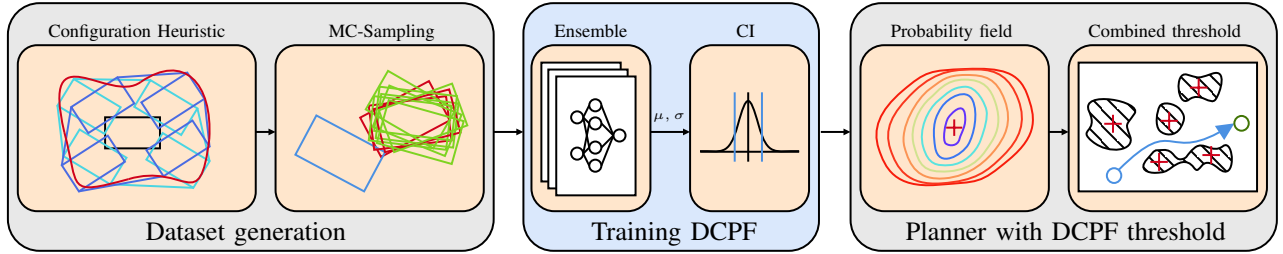}
    \captionof{figure}{Proposed pipeline for learning and planning with Deep Collision Probability Fields (DCPF): (\textbf{Left}) we start by generating a dataset with Monte Carlo sampling, balanced by selecting the configurations with a heuristic based on Minkowski Sums;  (\textbf{Centre}) an ensemble of DCPF networks is trained on this dataset. The upper bound of the Confidence Interval (CI) of this ensemble is then used to approximate collision probability fields for an obstacle. (\textbf{Right}) In a scenario with multiple obstacles, we can threshold the combined collision probability estimate to exclude unsafe states when planning.}\vspace{-0.15cm}
    \label{fig:main_figure}
        \end{center}
    }]
}


\maketitle
\thispagestyle{empty}
\pagestyle{empty}

\begin{abstract}
Estimating collision probabilities between robots and environmental obstacles or other moving agents is crucial to ensure safety during path planning. This is an important building block of modern planning algorithms in many application scenarios such as autonomous driving, where noisy sensors perceive obstacles.
While many approaches exist, they either provide too conservative estimates of the collision probabilities or are computationally intensive due to their sampling-based nature. 
To deal with these issues, we introduce Deep Collision Probability Fields, a neural-based approach for computing collision probabilities of arbitrary objects with arbitrary unimodal uncertainty distributions.
Our approach relegates the computationally intensive estimation of collision probabilities via sampling at the training step, allowing for fast neural network inference of the constraints during planning. In extensive experiments, we show that Deep Collision Probability Fields can produce reasonably accurate collision probabilities (up to $10^{-3}$) for planning and that our approach can be easily plugged into standard path planning approaches to plan safe paths on 2-D maps containing uncertain static and dynamic obstacles. Additional material, code, and videos are available at  \url{https://sites.google.com/view/ral-dcpf}.
\end{abstract}

\section{Introduction}
\label{sec:intro}
Safety is one of the most critical issues that needs to be solved to deploy autonomous agents in the real world. This is particularly important for autonomous robots and cars, as the perception of the world is subject to different sources of uncertainty such as noisy sensor measurements, approximate target tracking models, sensor malfunctioning, or adversarial attacks~\cite{banfi_path_2020}. To deal with uncertainty, path planning and control approaches could incorporate some form of probabilistic constraints~\cite{hardy_contingency_2013}. Using such constraints, we can force the \gls{cp} to take very small values while preventing the planner from generating overly conservative paths, leveraging the accurate uncertainty
estimate coming from modern perception systems.

However, even for 2D cases, with robots and obstacles of simple shapes (cf. Fig.~\ref{fig:main_figure}), whose pose is described by a Gaussian distribution, it is impossible to calculate the \gls{cp} in closed form.
To incorporate CP constraints in path planning, researchers explored two main directions: approximate methods, such as~\cite{bry2011rapidly,lee2013sigma,kamel2017robust,thomas2021exact}, and sampling-based methods~\cite{lambert_collision_2008,schmerling_evaluating_2017, banfi_path_2020}. 
Both approaches have some key issues. To ensure the satisfaction of the \gls{cp} constraints, approximate methods are prone to be over-conservative, preventing the robot from maneuvering in tight gaps. Conversely, sampling-based methods are often computationally intensive and do not scale well when reducing the \gls{cp} constraint. This issue is critical in autonomous driving since it requires real-time decisions and, at the same time, very low \gls{cp} values.

To tackle these issues, we introduce~\gls{dcpf}, a neural-based approach for computing collision probabilities. The key idea of \gls{dcpf} is to relegate a time-consuming Monte-Carlo estimate of the \gls{cp} to the dataset generation phase while allowing for fast neural network evaluation of the constraints during planning. Our network design is inspired by \gls{sdf}, a technique extensively used to quickly compute distances between objects of arbitrary shapes. We reformulate the \gls{sdf} approach in the setting of collision probabilities and, based on the ideas presented in~\cite{liu_regularized_2022}, we introduce a novel inductive bias that enforces reasonable predictions even where the training data is sparse or non-existent, by gradually reducing the \gls{cp} to zero as the distance between the two objects increases and forcing the \gls{cp} to approach one if the distance is small.
\gls{dcpf} has three significant advantages. First, we can learn the \gls{cp} from data under arbitrary probability densities. Second, we can create smooth and differentiable distance fields, a particularly desirable property for trajectory optimization. Finally, using deep neural networks allows for fast and parallel querying of data points, combining the strengths and avoiding the weaknesses of the approaches presented in the literature. 

In extensive experiments, we show that Deep Collision Probability
Fields can produce reasonably accurate collision probabilities (up
to $10^{-3}$) for planning, and we demonstrate the effectiveness of our approach in many different simulated path planning tasks under uncertainty, in terms of computational time and CP accuracy. We conclude the paper with validation in the real world in a navigation scenario with two TIAGo mobile robots.

\subsection{Related Work}
Prior works that studied CP constraints in the context of path planning can be classified along several orthogonal dimensions. The most important are (1) the method used to compute the CP, namely, sampling-based or approximated, (2) the type of uncertainty (Gaussian or more complex), (3) where uncertainty is assumed to be present (robot, obstacles, or both), and (4) whether the CP constraint is enforced on a single trajectory step, or along the full trajectory. Due to abundant literature on the subject, only a few approaches are discussed below. We refer to~\cite{thomas2021exact} for a more comprehensive overview.

Lambert et al.~\cite{lambert_collision_2008} assume Gaussian uncertainty on both the robot's and obstacles' pose, and use Monte Carlo sampling to compute step-wise collision probabilities which, in turn, allow to compute a safe speed profile. More recently, Schmerling and Pavone~\cite{schmerling_evaluating_2017} considered uncertainty in the robot's execution of its nominal trajectory, and use Monte Carlo with importance sampling to compute the trajectory CP. Banfi et al.~\cite{banfi_path_2020} consider complex, potentially multi-modal obstacles' uncertainties that might arise in adversarial contexts, and propose a method based on the Sequential Ratio Probability Test (SPRT) to compute CPs for partial trajectories obtained during planning. A common advantage of these sampling-based methods is that they naturally come with theoretical guarantees, as it is easy to identify precise standard errors, confidence intervals, etc., of the computed CPs. However, a common downside of these methods is that they are computationally intensive, and might require thousands of samples to give reasonably precise estimates of small CPs. While in this paper we leave the investigations of theoretical guarantees for future work, we remark that our approach could already be used to identify promising regions of the planning space on which such guarantees could be derived via sampling.

Other approaches for CP estimation consider different types of approximations to reduce the computation time. Bry and Roy~\cite{bry2011rapidly} assume the uncertainty on the robot's pose to be Gaussian and use the approximation of checking the ellipse defined by the covariance matrix and a desired chance bound for enforcing step-wise CP constraints. A similar approach is used by Kamel et al. in~\cite{kamel2017robust}. Hardy and Campbell~\cite{hardy_contingency_2013} consider Gaussian uncertainty on the obstacles' positions and use rectangular bounding boxes for both the robot and the obstacles to obtain a CP upper bound. Thomas et al.~\cite{thomas2021exact} consider Gaussian uncertainty on both the robot's pose and obstacles and approximate their shapes as ellipsoids. This allows us to formulate the collision condition as a distance between ellipsoids. These methods are very fast, but their conservative nature might prevent the robot from finding feasible paths that are both safe and low-cost.

\subsection{Problem Formulation}
We consider a mobile robot operating in a $n$-dimensional environment
specified by a bounded region $\planningspace\subset\mathcal{R}^n$.
We assume $X$ is composed of a static untraversable region $\staticobstaclespace\subset X$ ---known without uncertainty--- and free space $\freespace\subseteq X$, with $\freespace\cap \staticobstaclespace = \emptyset$. 
Let $O = \lbrace o^k\rbrace$ be a set of $k$ independent obstacles. Each obstacle is completely
described by the tuple \mbox{$o = \langle \objectconfigspace, \objectdensity, \objectshape\rangle$}
where $\objectconfigspace\subset \mathbb{R}^n$ is the space of all possible object configurations ---such as position, orientation, and lengths--- describing the properties of the obstacle, $\objectdensity$, is a probability distribution of each obstacle configuration $\objectconfig\in\objectconfigspace$, representing the uncertainty of the perceived obstacle, and $\objectshape$ is the set of points occupied by the obstacle assuming that the true obstacle configuration is $\objectconfig$. We assume that the robot's position is known without uncertainty. Therefore, the robot can be defined as the tuple \mbox{$r = \langle \robotconfigspace, \robotshape \rangle$},
where $\robotconfigspace$ denotes the configuration space of the robot and $\robotshape$ denotes the set of points occupied by the robot at configuration $\robotconfig\in\robotconfigspace$.
In this paper, we focus on the special case where the robots and the obstacles are 2D rectangles with sides $l_1$, and $l_2$, with isotropic Gaussian uncertainty. Under these assumptions, the obstacles can be described by the following parameter vector $\objectconfig=\left[x, y, \phi, l_1, l_2\right]^T$ with $\objectconfig\sim\mathcal{N}(\vmu, \mSigma)$.

The collision probability $p_{\text{coll}}(r, o)$ between robot $r$ and obstacle $o$ is defined as
\begin{equation}
    \scalemath{0.9}{p_{\text{coll}}(r, o) = \int_{\mathcal{D}}  \objectdensity d\objectconfig,}
    \label{eq:collision_probability}
\end{equation}
where $\mathcal{D} = \left\{ \vq_o \, | \, \objectconfig \in \objectconfigspace, \robotshape \cap \objectshape \neq \emptyset \right\}$.
Our goal is to compute an approximated version of $p_{\text{coll}}(r, o)$ with a neural network, which we denote $\hat{p}_{\text{coll}}(r, o)$, and show that it can be used in several planning settings. To this aim, we define the following family of planning problems.

Let $\robotconfig(0)$ be the initial robot configuration and $\goalset\subset\freespace$ is the set of admissible goals states.
Let $\mathcal{P}$ be the space of all possible paths. We assume that every path $\pi=\left[a_0, \dots, a_T\right]\in\mathcal{P}$ is composed by a set of $T$ motion primitives $a(t)$. Each motion primitive moves the robot from a configuration $\robotconfig(t)$ to a new configuration $\robotconfig(t + \Delta t)$. 
Finally, let $r(t)$ and $o(t)$ denote the tuples describing robot and obstacle at step $t$. 

We consider planning problems of the following form
\begin{align}
    \scalemath{0.9}{\argmin_\pi} & \scalemath{0.9}{\quad c(\pi) = \sum_{a_t \in \pi} c(a_t) \nonumber}\\
    \scalemath{0.9}{\text{s.t.}} & \scalemath{0.9}{\quad
    1 - \prod_i \left(1- p_\text{coll}(r(t), o_{i}(t)) \right) \leq p_{\max}\quad  \forall t, \nonumber}\\
    & \scalemath{0.9}{\quad \robotconfig(t) \in \freespace\, \forall t \nonumber, \quad \robotconfig(T) \in \goalset,}
\end{align}
where $p_{\max}\in[0,1]$ is the maximum \gls{cp} allowed for the planned trajectory and $c(\pi)$ is a cost function, like path length, computed by summing the cost of each motion primitive composing the path.

Notice that in this formulation we assume that the robot will be able to perfectly track the computed trajectory, but we do not make any assumption regarding the number of obstacles, their nature--- static or dynamic ---and the magnitude of their future uncertainty. For dynamic obstacles tracked with a Kalman filter, this could simply be obtained by applying the filter prediction step~\cite{hardy_contingency_2013}.

\section{Deep Collision Probability Fields}
\label{sec:methodology}
Our goal is to approximate the collision probability between a robot and an arbitrary object efficiently by means of a neural network, hence, we introduce \gls{dcpf} that exploit neural approximations to obtain rapidly accurate \gls{cp} estimates.
We make two modeling assumptions: first, we consider the coordinate frame defined in the obstacle's center of mass, instead of the world coordinate frame. Second, we assume a parametric form for the obstacle's probability distribution, $\parametricobjectdensity$ where $\densityparamsobj$ represents the distribution's parameters. For example, assuming Gaussian uncertainty, the distribution parameters are the covariance matrix entries, $\densityparamsobj = \mSigma_o$.  Notice that, given that the coordinate system is obstacle-centric, the obstacle distribution mean for position and rotation is zero. We assume that the distribution family is known and that the distribution parameters are estimated from a perception system. 
Thanks to these assumptions we can easily impose an inductive bias in the network's structure, forcing the probability of collision to smoothly decrease to zero, when the distance between the object and the robot increases, while making it close to one when the two objects overlap.
This inductive bias, inspired by the one presented in~\cite{liu_regularized_2022}, exploits the fact that sufficiently far from the object, both the shape and the object uncertainty are irrelevant, as the collision probability will drop to zero. This bias allows us to efficiently learn the probability field for any configuration while keeping the probability approximation smooth.

\subsection{Network structure}
Using the above-mentioned assumptions and implementing the distance-based inductive bias, we obtain the following network structure
\begin{align}
    \scalemath{0.9}{\hat{p}_\text{coll}(\robotconfig, \densityparamsobj) =}& 
    \scalemath{0.9}{(1-\sigma^1_{\vtheta}(\robotconfig, \densityparamsobj))\left((1-\sigma^2_{\vtheta}(\robotconfig, \densityparamsobj)) f_{\vtheta}(\robotconfig, \densityparamsobj)\right)} \nonumber\\
    &\scalemath{0.9}{+ \sigma^1_{\vtheta}(\robotconfig, \densityparamsobj),}
\end{align}
with the current robot configuration $\robotconfig$, the  parameters of the obstacle density $\densityparamsobj$, and
the vector of learnable parameters of the neural network $\vtheta$.
Furthermore, for $i\in\{1,2\}$, we define two functions as 
\begin{equation}
\scalemath{0.9}{\sigma^i_{\vtheta}(\robotconfig, \densityparamsobj) = \text{sigmoid}\left(s^i\alpha^i_{\vtheta}(\robotconfig, \densityparamsobj)\cdot\left(\rho^i_{\vtheta}(\robotconfig, \densityparamsobj)-||\vx||_2 \right)\right),}
\end{equation}
where $\rho^i_\vtheta(\robotconfig, \densityparamsobj)$ defines a soft threshold for switching from a local approximated collision probability to the one predicted by the inductive bias (zero or one), $\alpha^i_\vtheta(\robotconfig, \densityparamsobj)$ regulates the sharpness of the change between the two modes, and $s^i$ is a sign multiplier, with $s^1=1$ and $s^2=-1$. The first regularizer $\sigma^1_{\vtheta}$ biases the output towards one when the query point is close to the obstacle.  The second regularizer $\sigma^2_{\vtheta}$ moves the value of $p_{coll}$ closer to 0 when the robot is close to the obstacle.

\begin{figure}[tb]
\centering
\input{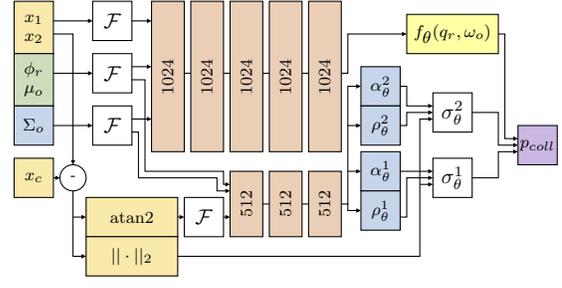}
\caption{The network structure of \gls{dcpf}: input data is processed using Fourier features. The processed features are fed to a deep neural network with 5 fully connected 1024-dimensional hidden layers. 
An additional set of 3 fully connected 512-dimensional hidden layers compute the input for the shaping functions $\alpha$ and $\rho$, which influence the mode switching of the two regularizers $\sigma_\vtheta^1$ and $\sigma_\vtheta^2$, that implement the euclidean distance bias.}
\label{fig:architecture}
\vspace{-2em}
\end{figure}
The architecture of \gls{dcpf}, depicted in Fig. \ref{fig:architecture}, is composed of two neural networks. The larger network consists of 5 fully connected hidden layers, each 1024-dimensional. This network takes the robot's position and configuration, and the obstacle's configuration and variance, each individually encoded with random Fourier features to compute $f_\theta$. 
In addition, the smaller network uses 3 fully connected 512-dimensional hidden layers to compute $\alpha_\theta^{i}$ and $\rho_\theta^{i}$, the shaping parameters for the two regularizers.
As input of this network, we exchange the positional information of the robot with just the angle from the position in polar coordinates, as the distance to the center $||\vx||_2$ is directly used in the regularizer.

We use GeLU activations for all layers except for the output layers of the small network, which have their specific activation functions.
In particular, we limit the range of $\alpha_\vtheta^{1}$ and $\alpha_\vtheta^{2}$ to the interval $[1,21]$ by applying a sigmoid function and a bias on the output unit. $\rho_\vtheta^{1}$ is bounded to the interval $[0, 12]$ and $\rho_\vtheta^{(2)}$ is constrained to be positive, as the last layer is a softplus. The ranges discussed above are selected to give reasonable output values and ensure the stability of the training process in the autonomous driving setting.  We also constrain the collision probability output $f_{\vtheta}$ in the interval $[0,1]$, using a sigmoid function.

\paragraph*{Conservative \gls{cp} estimates} 
Approximating a function with a neural network may lead to approximation errors, especially in areas far from the original dataset distribution. 
To reduce the chances of significant approximation errors, we learn an ensemble of \gls{dcpf}s. Using this approach, we are able not only to provide a more accurate value of the collision probability but also to obtain a measure of the model prediction uncertainty.  
In the rest of our work, on top of the vanilla approach using a single neural network, we will exploit the ensemble technique in three different ways. The first option is to take the \textit{mean}, where we average the prediction of multiple ensembles, reducing the error due to overfitting. Alternatively, we can take the \textit{maximum} collision probability value, ensuring that the estimate we select is always conservative. This is particularly useful in case the collision probability constraint bound is critical. Finally, we can exploit the measure of the prediction uncertainty by computing the 95\% confidence interval of our prediction to build an upper confidence bound around the mean.

\subsection{Data generation}
\label{sec:data-gen}
We resort to a sampling-based method to generate an accurate dataset that does not leverage approximations to compute CPs. Specifically, since we are not bound by time constraints at this stage, we resort to Simple Monte Carlo sampling. 
Another key observation is that the desired accuracy of the \gls{cp} estimate directly depends on the \gls{cp} itself. For example, suppose that after having drawn 10k samples, our current \gls{cp} estimate is .5; in this case, we would accept an error of .01, especially if our planning constraint $p_{\text{max}}$  is .01 or .1. Following from this observation, we define the following probability intervals: $[0, .01)$, $[.01, .1)$, $[.1, 1]$. Each interval is then associated with a different desired accuracy for a \gls{clt} based 95\% confidence interval estimate\footnote{See~\cite{mcbook}, Eq. (2.20), and the following paragraph for the special case of estimated \gls{cp} = 0 or 1.}: $\pm.01$, $\pm.001$, $\pm10^{-4}$. Therefore, we can stop drawing samples as soon as the current \gls{clt}-based 95\% confidence interval falls below the accuracy associated with the current \gls{cp} estimate. To speed up this procedure, we only recompute the \gls{cp} estimate and associated confidence interval after having drawn a batch of samples. We use batches of $4\cdot 10^4$ samples, and $4\cdot 10^6$ maximum total samples, computed as the worst-case number of samples needed for the smallest probability interval. 

Another issue to consider during the training is to sample properly the robot configurations to be evaluated. Indeed, if the dataset size is a concern, sampling uniformly may require a prohibitively large amount of samples to properly cover the areas where the estimation of collision probability is particularly sensible. 
We propose a sampling method based on the \gls{minsum} between the robot and the obstacle. The \gls{minsum} is the shape obtained by sliding the robot shape around the obstacle~\cite{claes_collision_2012}. 
Our key idea is to compute a shape that resembles the isolines of the \gls{cp} landscape, particularly close to the constraint budget level, and subsequently sample points from it.
In a scenario without uncertainty, this shape can be trivially obtained by computing the \gls{minsum} between the robot and the obstacle. To take the uncertainties $\mSigma_o=(\sigma_x,\sigma_y,\sigma_{\phi},\sigma_{l_1},\sigma_{l_2})$ into account we apply various strategies. 
For positional and shape variance we construct an inflated shape $s_\text{inflated}$ by computing a \gls{minsum} between the shape of the obstacle $s_o$ and an ellipse $s_e \sim N(\sigma_x + \sigma_{l_1}, \sigma_y +\sigma_{l_2} | \left(\sigma_x + \sigma_{l_1}, \sigma_y +\sigma_{l_2}\right))$.
The effects of rotational variance are approximated by constructing the shape $s_\text{rotational}$, by computing the union of multiple $s_\text{inflated}$ for rotated obstacles $s_o$. On top of that, we smooth the obtained shape by growing and shrinking $s_\text{rotational}$. The rotations are a discretization of the interval  $[-\phi_m, \phi_m]$ with $\phi_m \sim \mathcal{U}(\sigma_\theta, 3.1\cdot\sigma_\theta)$.
In our experiment, we found that three rotations are sufficient, for this heuristic to result in sample configurations for a well-balanced dataset.

\subsection{Training} 
We train our model by optimizing the following loss
\begin{equation}
    \scalemath{0.9}{\mathcal{L}(\mathcal{D}) = \sum_{\robotconfig, \densityparamsobj, \bar{p}\in\mathcal{D}}\mathcal{H}\left(\hat{p}_{\text{coll}}\left(\robotconfig, \densityparamsobj\right), \bar{p}\right)
    +\gamma \cdot \mathcal{R}\left(\robotconfig, \densityparamsobj\right),}
\end{equation}
where $\mathcal{H}$ is the binary cross-entropy between the \gls{cp} prediction of the network, $\bar{p}$ the target value computed in the dataset, and $\mathcal{R}$ is a regularization term, weighted by a  $\gamma$ coefficient, composed by the sum of the following two terms 
\begin{equation}
    \scalemath{0.9}{\mathcal{R}\left(\robotconfig, \densityparamsobj\right) = 
    \left|\Delta \rho_\theta\right| +
    \sum_{i=\{1,2\}}
    \text{sigmoid}\left(
    s^i\alpha_\theta^i\cdot\Delta\rho_\theta/2
    \right),}
\end{equation}
with $\Delta\rho_\theta=\rho^2_{\boldsymbol \theta}-\rho^1_{\boldsymbol \theta}$ the difference of the mode switching parameters. 
The first term of the regularizer forces the network to switch between the regimes of data-driven collision probabilities to the inductive bias of zero or one probabilities as soon as possible, by bringing the thresholds $\rho^i_\vtheta$ closer together. 
The second term enforces the thresholds to be in the right order by minimizing the influence of the bias in their center. The regularization term is particularly important in areas with low sample density. 

\input{sections/experiments}

\section{Conclusion}
\label{sec:conclusion}
In this paper, we introduced \gls{dcpf}, a neural approach to compute efficiently \gls{cp}. Differently from sampling-based approaches, \gls{dcpf} shifts the heavy part of the computation offline, allowing for a fast, time-consistent, yet very accurate, evaluation of \gls{cp} during planning. 
Our method can easily incorporate obstacle uncertainty in many different settings, including dynamic environments and shape uncertainty. 
Furthermore, \gls{dcpf} provides a differentiable representation of the collision probability, allowing for an easy combination with trajectory combination methods.
Additionally, to increase the applicability in safety-critical settings, we present an ensemble approach to deal with model prediction uncertainty and provide conservative \gls{cp} estimations.

Our experiments show that \gls{dcpf} can speed up the \gls{cp} evaluation, particularly in the setting of parallelizable planning algorithms. As the inference time of the neural network is consistent, we are free from the need to specify a planning time budget, which may cause the standard method to wrongly label safe configurations. Our planning experiments show that \gls{dcpf} provides a better evaluation of the safety of the states than the sampling-based methods with a limited computation budget.
Finally, we show that we can deploy our approach in the real world, by performing a real-world overtaking task using two Tiago robots, under perception and dynamic uncertainty.
While we focus on the simple setting of rectangular obstacles (crucial for autonomous driving), \gls{dcpf} can be easily extended to support arbitrary shapes, as it only requires generating a proper dataset. This includes non-convex shapes that are particularly computationally intensive for sample-based \gls{cp} estimation methods. 

In future works, we will investigate the applicability of our approach in more challenging settings and we will combine our method with state-of-the-art parallel planning algorithms: this would allow us to reduce massively the planning time, allowing methods of planning under uncertainty to scale to complex real-world scenarios. 





\section*{ACKNOWLEDGMENT}
This researcher has been supported by the BMBF collaborative project KIARA (grant no. 13N16274), and the the DFG EN Project iROSA (CH 2676/1-1).
The authors acknowledge the use of the high-performance computer Lichtenberg at the NHR Centers NHR4CES at TU Darmstadt. This is funded by the Federal Ministry of Education and Research, and the state governments participating based on the resolutions of the GWK for national high-performance computing at universities.

We sincerely thank  Sophie L\"uth for her help and technical assistance with the real-world deployment, and Matteo Luperto for providing evaluation datasets.


\bibliographystyle{IEEEtran}
\bibliography{bibliography}

\clearpage

\input{sections/appendix}

\end{document}

%% file: commands/preamble.tex
\usepackage{amsmath} 
\usepackage{amssymb} 
\usepackage{bm}
\usepackage{graphicx}
\usepackage{glossaries}
\usepackage{multirow} 
\usepackage{tikz}
\usepackage{pgfplots}
\usepackage{pgfplotstable}
\usepackage{url}
\usepackage{booktabs}
\usepackage{siunitx}

\usepackage{silence}
\usepackage{caption}

\DeclareCaptionLabelSeparator{periodspace}{.\quad}

%% file: commands/tikz.tex
\pgfplotsset{compat=newest}
\usepgfplotslibrary{statistics}
\usepgfplotslibrary{colorbrewer}
\usetikzlibrary{patterns}

\usetikzlibrary{pgfplots.groupplots}

\pgfmathdeclarefunction{fpumod}{2}{%
        \pgfmathfloatdivide{#1}{#2}%
        \pgfmathfloatint{\pgfmathresult}%
        \pgfmathfloatmultiply{\pgfmathresult}{#2}%
        \pgfmathfloatsubtract{#1}{\pgfmathresult}%
        \pgfmathfloatifapproxequalrel{\pgfmathresult}{#2}{\def\pgfmathresult{5}}{}%
    }

\pgfmathdeclarefunction{myCustomFunction}{1}{%
    \pgfmathparse{
    #1 > 5 ? 
        ( 1/7 + 2 + 1/7*fpumod(#1, 6) ) :
        (#1 > 3 ?
            (1/4 + 1 +  1/4*fpumod(#1 - 3, 3)) :
            (floor(#1/3) + 1/4*fpumod(#1,3) + 1/4 )
        )
    }%
}

\pgfplotscreateplotcyclelist{mycolorlist}{
    {Greens-G},
    {Greens-I},
    {Greens-K},
    {Blues-G},
    {Blues-I},
    {Blues-K},
    {Reds-D},
    {Reds-F},
    {Reds-G},
    {Reds-I},
    {Reds-K},
    {Reds-M},
    {Reds-O}
}

\pgfplotsset{
    /pgfplots/custom legend/.style={
        legend image code/.code={
            \draw [only marks,mark=diamond*,mark options={scale=1}]
            plot coordinates { 
                (0.3cm,0cm)
            };
        }, 
    },
}

\pgfplotsset{
    boxplot prepared from table/.code={
        \def\tikz@plot@handler{\pgfplotsplothandlerboxplotprepared}%
        \pgfplotsset{
            /pgfplots/boxplot prepared from table/.cd,
            #1,
        }
    },
    /pgfplots/boxplot prepared from table/.cd,
        table/.code={\pgfplotstablecopy{#1}\to\boxplot@datatable},
        row/.initial=0,
        make style readable from table/.style={
            #1/.code={
                \pgfplotstablegetelem{\pgfkeysvalueof{/pgfplots/boxplot prepared from table/row}}{##1}\of\boxplot@datatable
                \pgfplotsset{boxplot/#1/.expand once={\pgfplotsretval}}
            }
        },
        make style readable from table=lower whisker,
        make style readable from table=upper whisker,
        make style readable from table=lower quartile,
        make style readable from table=upper quartile,
        make style readable from table=median,
        make style readable from table=lower notch,
        make style readable from table=upper notch,
        make style readable from table=average
}

%% file: commands/acronyms.tex
\newacronym{cp}{CP}{Collision Probability}
\newacronym{dcpf}{DCPF}{Deep Collision Probability Fields}
\newacronym{sprt}{SPRT}{Sequential Probability Ratio Test}

\newacronym{sdf}{SDF}{Signed Distance Function}
\newacronym{tsdf}{TSDF}{Truncated Signed Distance Function}
\newacronym{esdf}{ESDF}{Euclidean Signed Distance Function}
\newacronym{redsdf}{ReDSDF}{Regularized Deep Signed Distance Fields}
\newacronym{apf}{APF}{Artificial Potential Fields}
\newacronym{rmp}{RMP}{Riemannian Motion Policies}
\newacronym{hri}{HRI}{Human-Robot Interaction}
\newacronym{ecomann}{ECoMaNN}{Equality Constraint Manifold Neural Network}
\newacronym{smpl}{SMPL}{A Skinned Multi-Person Linear Model}
\newacronym{poi}{PoI}{Points of Interest}
\newacronym{wbc}{WBC}{whole-body control}

\newacronym{sat}{SAT}{Separating Axis Theorem}
\newacronym{clt}{CLT}{Central Limit Theorem}

\newacronym{minsum}{MS}{Minkowski Sum}

\newacronym{mae}{MAE}{Mean Absolute Error}
\newacronym{rmsre}{RMSRE}{Rooted Mean Squared Relative Error}
\newacronym{pap}{PAP}{Percentage of Accurate Predictions}

%% file: commands/math.tex
\DeclareMathOperator*{\argmin}{arg\,min}

\DeclareMathOperator{\atantwo}{atan2}

\newcommand\scalemath[2]{\scalebox{#1}{\mbox{\ensuremath{\displaystyle #2}}}}

\def\vq{{\bm{q}}}

\def\vx{{\bm{x}}}

\def\vmu{{\bm{\mu}}}
\def\vtheta{{\bm{\theta}}}

\def\vomega{{\bm{\omega}}}

\def\mSigma{{\bm{\Sigma}}}

%% file: commands/notation.tex
\newcommand{\planningspace}{{X}}
\newcommand{\freespace}{\planningspace_{\text{free}}}
\newcommand{\staticobstaclespace}{\planningspace_{\text{obs}}}

\newcommand{\objectconfigspace}{{\mathcal{C}_o}}
\newcommand{\objectconfig}{{\vq_o}}
\newcommand{\objectshape}{{\mathcal{V}_o(\objectconfig)}}
\newcommand{\objectdensity}{{p_o(\objectconfig)}}
\newcommand{\densityparamsobj}{{\vomega_o}}
\newcommand{\parametricobjectdensity}{{p_o(\objectconfig;\densityparamsobj)}}

\newcommand{\robotconfigspace}{{\mathcal{C}_r}}
\newcommand{\robotconfig}{{\vq_r}}
\newcommand{\robotshape}{{\mathcal{V}_r(\robotconfig)}}

\newcommand{\goalset}{\mathcal{Q}_{G}}

%% file: sections/experiments.tex
\section{Evaluation}
We empirically evaluate \gls{dcpf} in three sets of experiments. The first experiment evaluates the \gls{cp} predictions obtained by \gls{dcpf} on a dataset inspired by an autonomous driving scenario, examining the impact of number and size of network layers. In the second experiment, we use the best network obtained in the first experiment to tackle two simulated planning scenarios: one with static obstacles and one with dynamic obstacles. In the third experiment, we validate \gls{dcpf} in the real world with two TIAGo robots (Fig.~\ref{fig:TIAGo}), where one acts as the agent and the other acts as a dynamic obstacle.

\begin{table}[b]
    \centering
    \vspace{-1em}
    \caption{Percentage of predictions within confidence interval table for different Network Sizes}
    \input{tables/network_ablation_ci_percentage}

    \label{table:network_ablation}
\end{table}

\begin{table}[b]
    \centering
    \vspace{-0.3cm}
    \caption{Inference computation time per sample}
    \input{tables/execution_speeds_reciproc}

    \label{table:inference_time}
\end{table}

\begin{figure}[t]
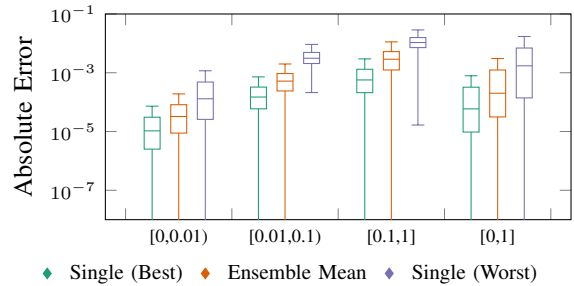

    \centering
    \input{img/experiments/01_network_evaluation/absolute_error_data}
    \input{img/experiments/01_network_evaluation/absolute_error_boxplot}

    \ref{abserrlegend}
    \caption{Box plot of absolute error evaluated on test dataset}
    \label{fig:binrrmse_box}
    \vspace{-0.5cm}
\end{figure}

\subsection{\gls{dcpf} Estimation of \gls{cp}}
We build a dataset of one billion samples, with an 80\%-10\%-10\% training-validation-test split, ensuring an equal balancing across the different probability intervals (see Section \ref{sec:data-gen}). We assume the robot dimensions to be fixed to a width of 4.07 and height of 1.74, while the $\objectconfig$ is obtained by drawing from a Gaussian distribution with $\mSigma_o \in [0,\sqrt 2]^5$. The dataset generation phase took 80 hours on a computer equipped with an AMD Ryzen 9 5950X 16-Core and an RTX 3080Ti (12 GB). We use the Adam optimizer for training with $2.4\cdot 10^{-4}$ learning rate. We evaluate the network varying the number of hidden layers in 3 to 7 and the number of hidden layers neurons in 128, 512, and 1024, while the network for the regularized was kept fixed to 3 layers of 512 neurons. For every setting, we use an ensemble of 10 networks and we evaluate the average prediction.
We consider two metrics, the \gls{mae} and the \gls{pap}. 
The \gls{mae} measures the absolute average error of the network prediction over the full dataset. Since our dataset samples are approximations we also use \gls{pap} to measure the percentage of accurate predictions, where we consider a prediction as accurate if it is within the confidence interval of the collision probability estimate.

In Table~\ref{table:network_ablation}, we report the ablation study on the network size using the \gls{pap} as metric. We obtain accurate predictions for each network size and the network precision improves increasing the number of neurons. However, increasing the number of layers does not have a consistent positive impact. Due to the results of our ablation, we select a network with 6 hidden layers of 1024 neurons for the subsequent experiments.

In Figure~\ref{fig:binrrmse_box}, we evaluate the performance of our selected network by reporting the boxplots of absolute errors for different \gls{cp} buckets. The results clearly show that our prediction error for the best network is very accurate, falling most of the time in the desired bucket accuracy. By looking at the worst prediction of each ensemble, we notice a significant drop in the accuracy. However, the presence of these outliers in the prediction is strongly mitigated by using the ensemble mean, which brings the distribution close to the best network, indicating that these errors are not very frequent in the dataset. This proves that, by using an ensemble technique, we provide highly accurate \gls{cp} computations, ensuring the safety of the planning algorithm. It is possible to enforce this safety guarantee with high probability by using a confidence interval to provide a conservative estimate of the network.

To evaluate the performance in terms of computation time, we compare our approach with the \gls{sprt} and the Z-Test methods~\cite{banfi_path_2020} to ensure the \gls{cp} is below a set of fixed constraints. We use $p_{\max}\in\{10^{-1}\,10^{-2},10^{-3}\}$ and a sample function based on the separating axis theorem. \textcolor{black}{The \gls{cp} distribution is approximately reciprocal.}
For every approach, we test the performance evaluation on the CPU using a single configuration, while for \gls{dcpf} we also evaluate the performance with batch sizes 1, 16, and 1024, in GPU and CPU. The computation time reported is the computation time per configuration, i.e., the total computation time divided by the batch size. Both Z-Test and \gls{sprt} sample a maximum of $4\cdot 10^6$ times for each constraint respectively. 
Our results, presented in 
Table~\ref{table:inference_time}, show that \gls{sprt} is a competitive method only when using a high \gls{cp} collision probability, and together with the Z-Test provides a highly variant behavior in terms of computation time. Our approach has much more consistent computation times. It is worth noting that the mean computation time can be heavily affected by the outliers, particularly for lower \gls{cp} budgets. Our method shines particularly with high batch sizes, making it suitable when checking collision probabilities of long trajectories or in combination with parallelized planners, e.g.~\cite{lambert2020stein,le2023accelerating,carvalho2023motion}.

\subsection{Planning with \gls{dcpf}}

\paragraph{Static Obstacles} For this work we consider two settings. A a narrow passage scenario with two obstacles and a random obstacle planning setting.

\begin{figure}[t]
    \centering
    \setlength{\tabcolsep}{1pt}
    \begin{tabular}{p{1.5em}ccc}
    & $p_{\max} = 10^{-1}$ & $p_{\max} = 10^{-2}$ & $p_{\max} = 10^{-3}$  \\
    \\
    \raisebox{3\normalbaselineskip}[0pt][0pt]{\rotatebox[origin=c]{90}{DCPF}} &
    \includegraphics[width=0.145\textwidth]{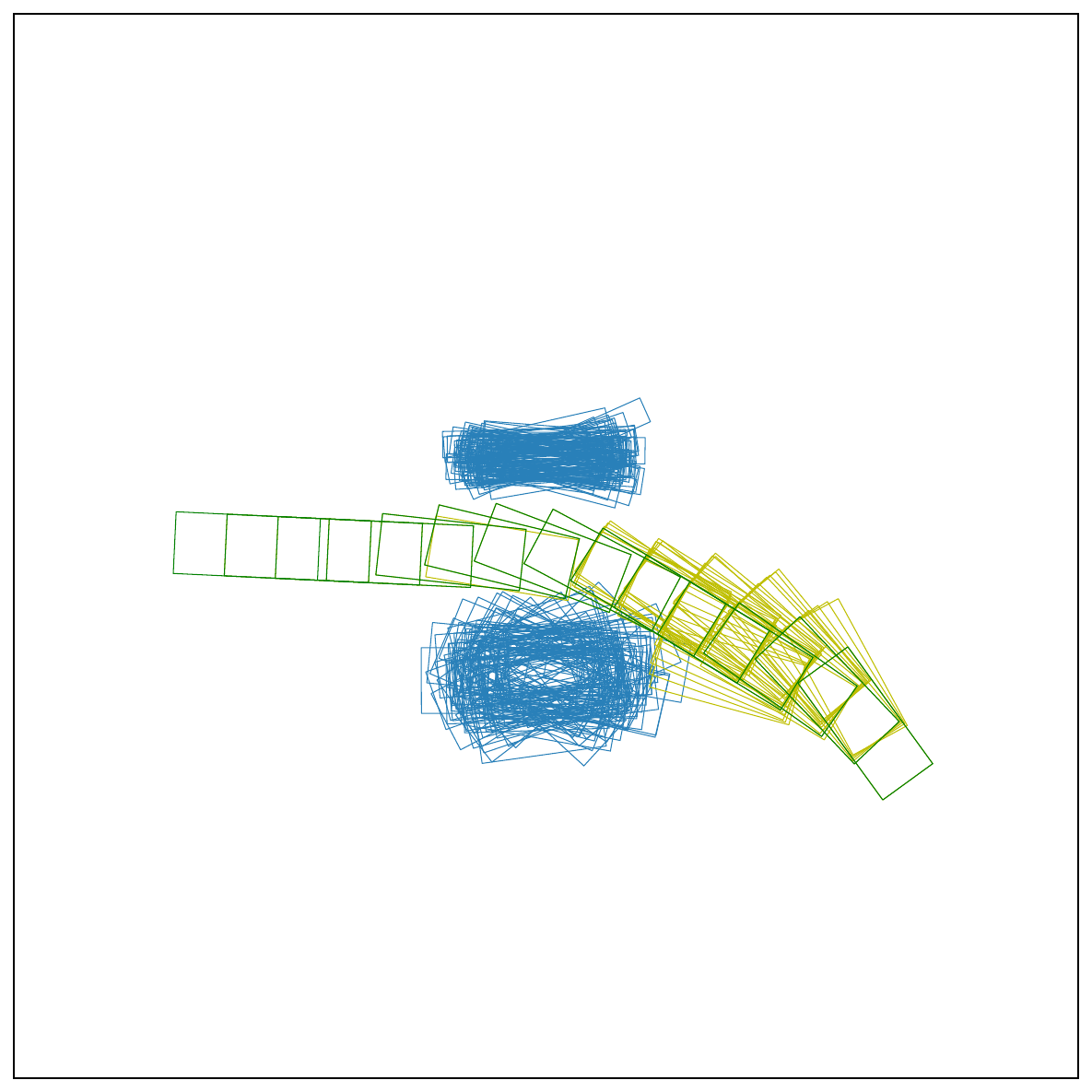} &
    \includegraphics[width=0.145\textwidth]{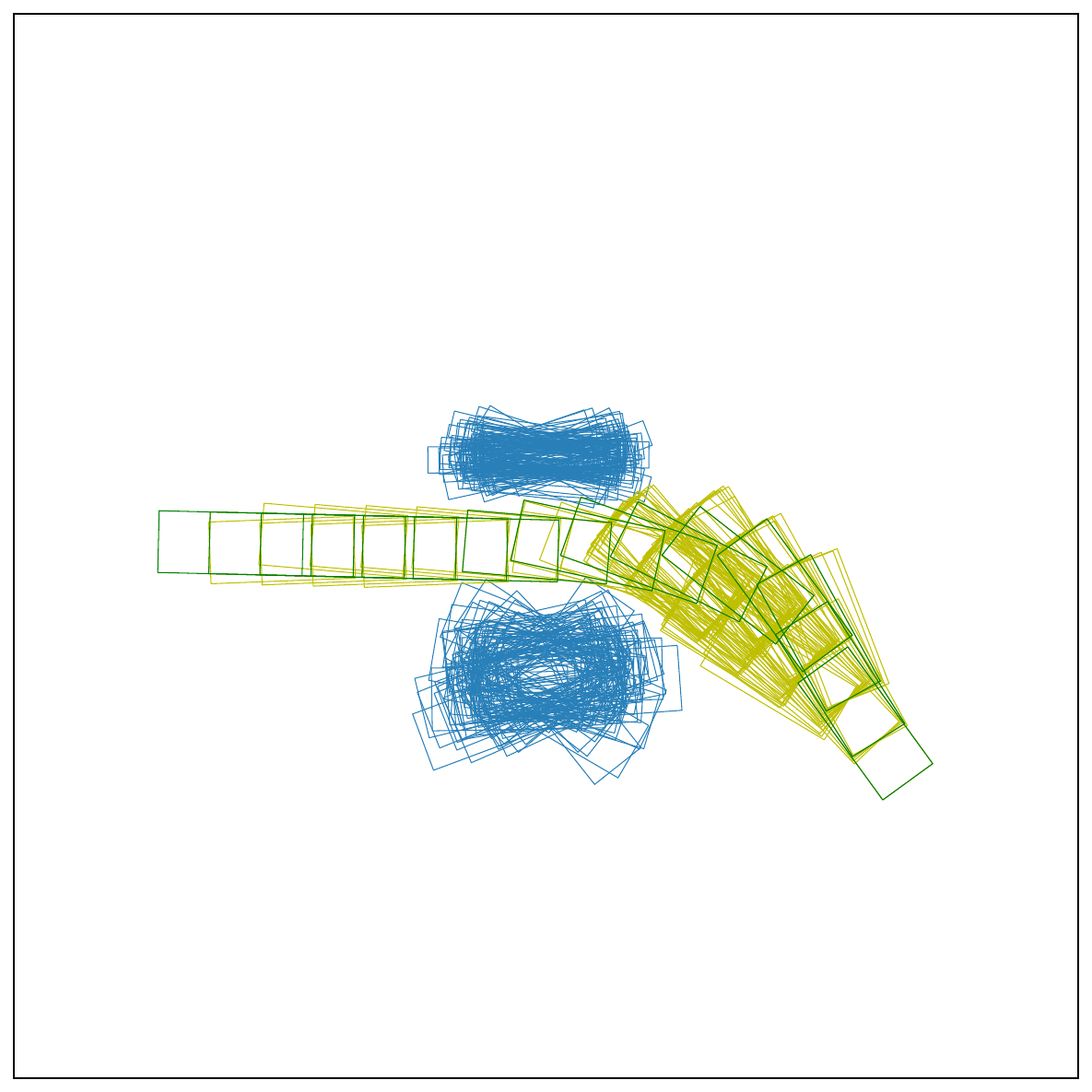} &
    \includegraphics[width=0.145\textwidth]{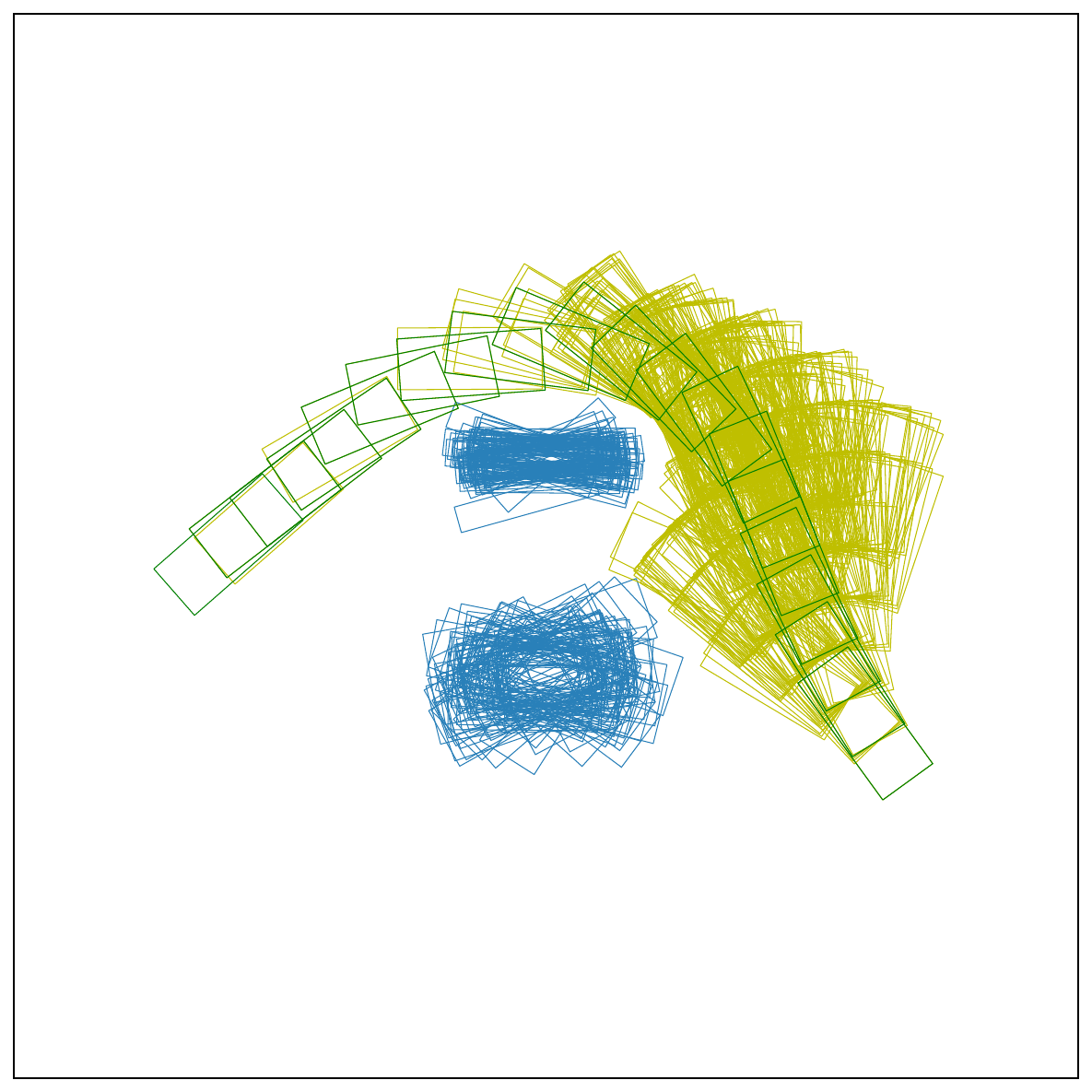} 
    \\
    \raisebox{3\normalbaselineskip}[0pt][0pt]{\rotatebox[origin=c]{90}{SPRT}} & 
    \includegraphics[width=0.145\textwidth]{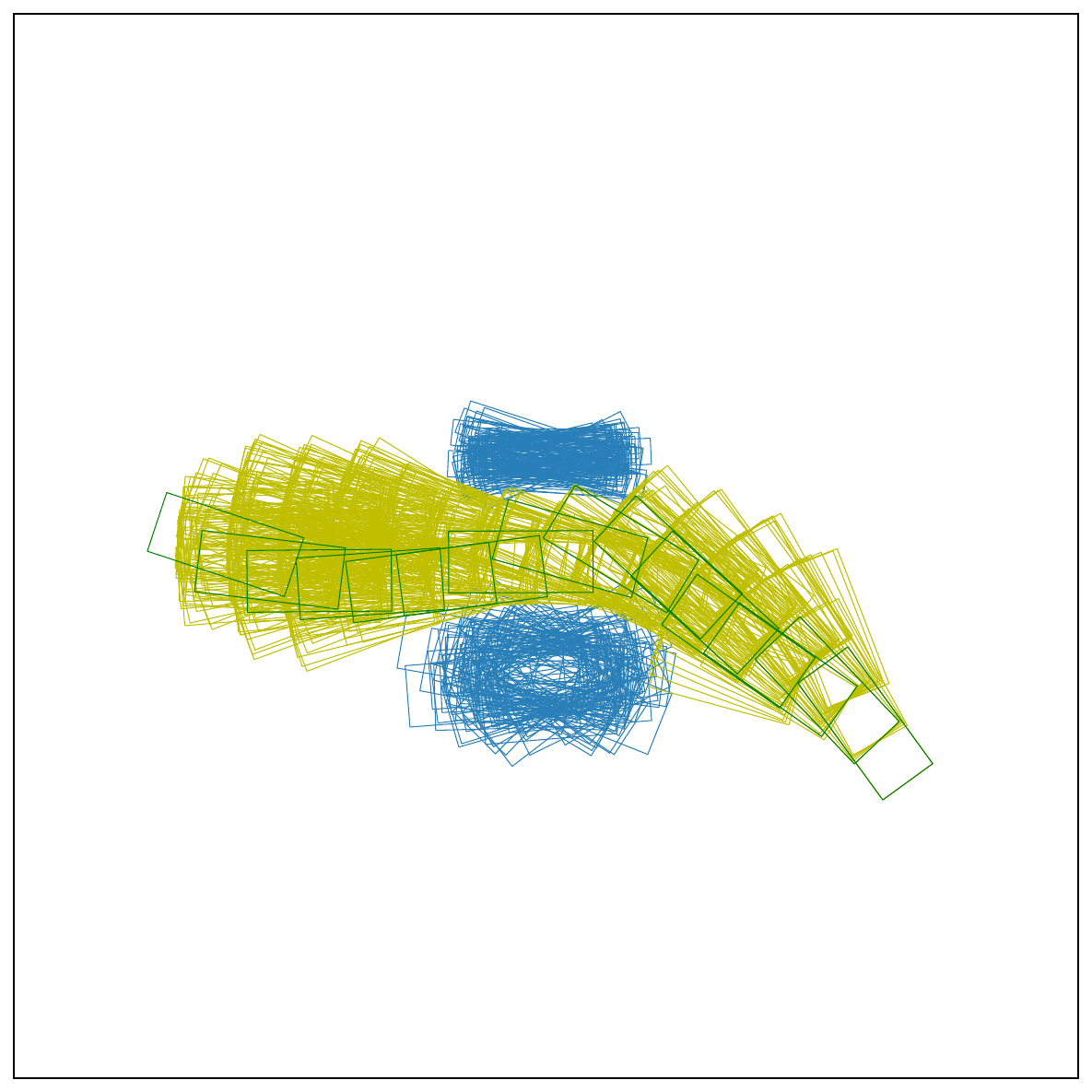} &
    \includegraphics[width=0.145\textwidth]{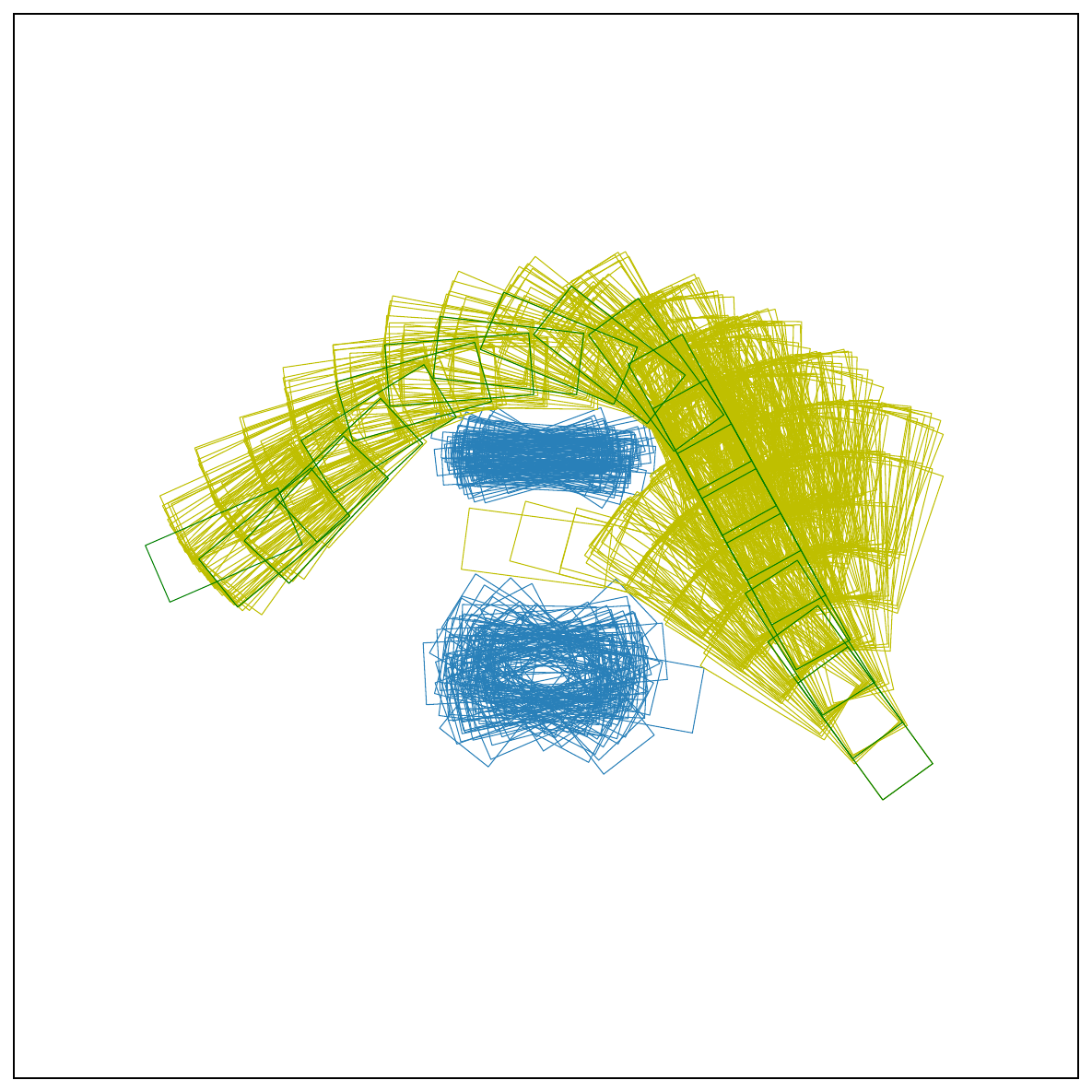} &
    \includegraphics[width=0.145\textwidth]{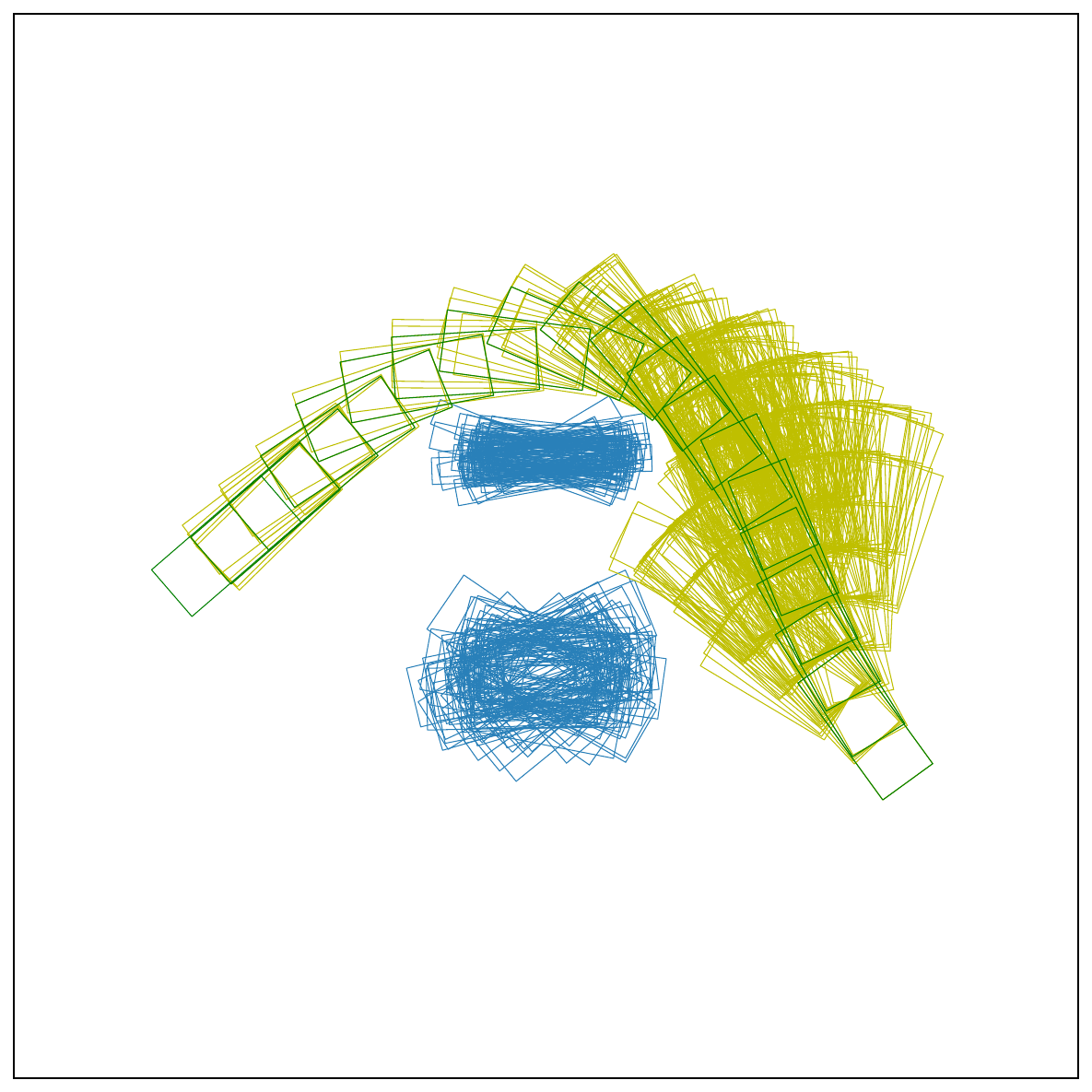} 
    \\
    \raisebox{3\normalbaselineskip}[0pt][0pt]{\rotatebox[origin=c]{90}{Z-Test}} & 
    \includegraphics[width=0.145\textwidth]{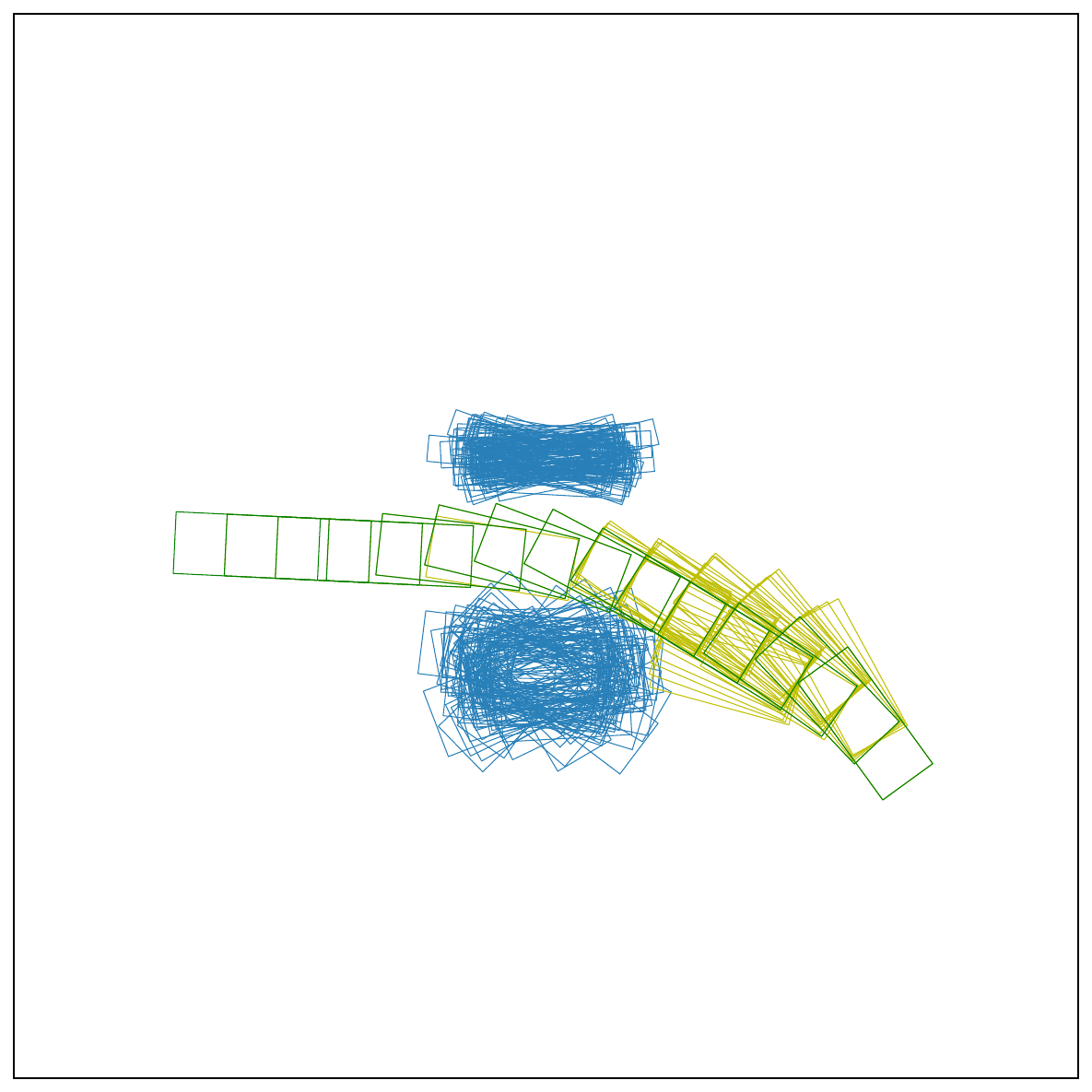} &
    \includegraphics[width=0.145\textwidth]{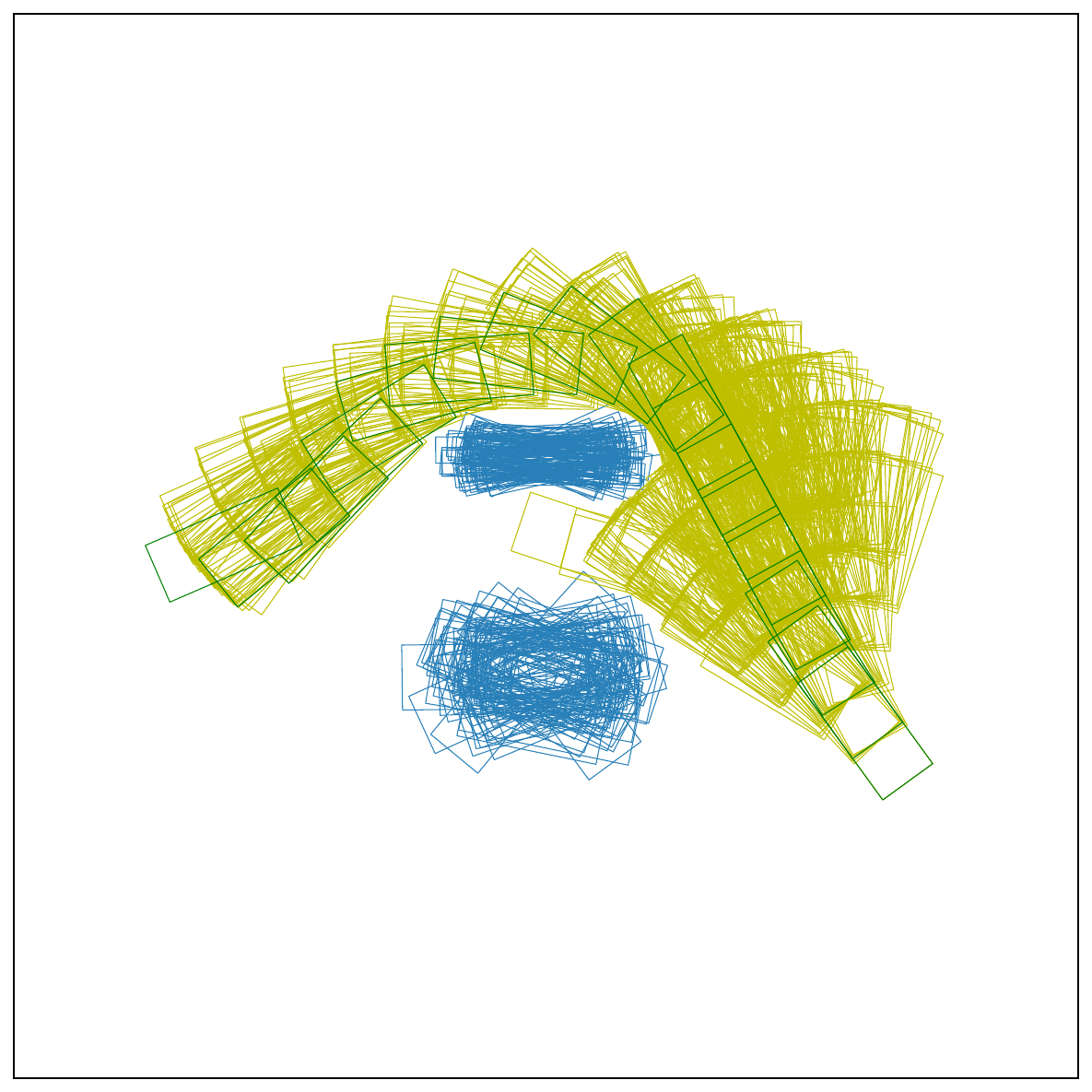} &
    \includegraphics[width=0.145\textwidth]{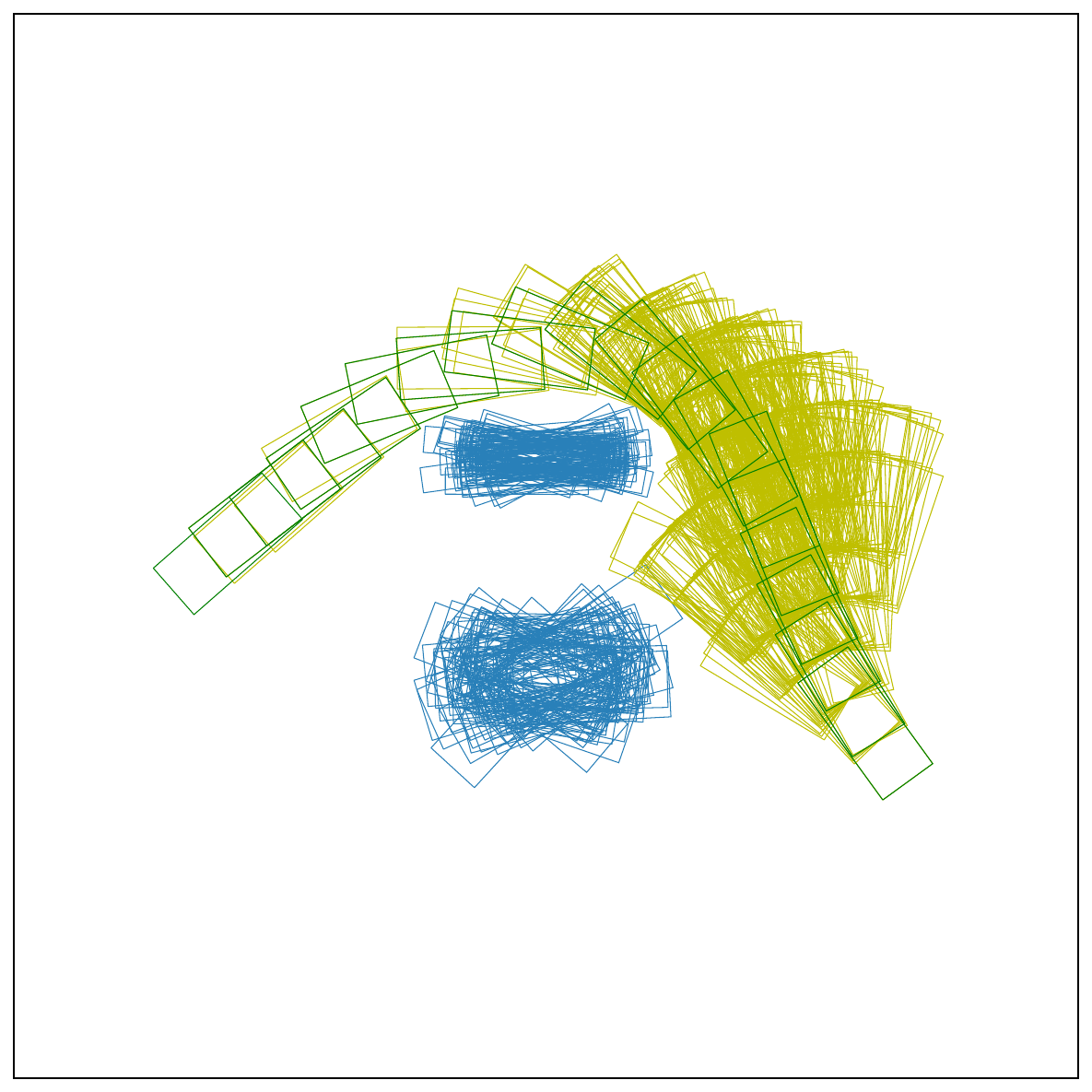} 
    \end{tabular}
    \caption{Comparison of paths generated using A* using different sampling algorithms. Blue rectangles represent obstacles sampled from the obstacle distribution, while the other rectangles represent a configuration of the robot checked by the planner. The configurations chosen as part of the solution are marked in dark green. \gls{sprt} samples a maximum of $4\times10^6$ and the Z-Test uses $10^5$ samples at most.}
    \label{fig:a_star_paths}
    \vspace{-2em}
\end{figure}

In the first planning scenario, we consider a square workspace containing two static obstacles forming a narrow passage, as shown Fig.~\ref{fig:a_star_paths}, with uncertainty described by 
\begin{align*}
\scalemath{0.9}{\mSigma_1} &= \scalemath{0.9}{\mathrm{diag}(0.05, 0.2, 0.03, 0.0001, 0.0001)} \\
\scalemath{0.9}{\mSigma_2} &= \scalemath{0.9}{\mathrm{diag}(0.15, 0.4, 0.13, 0.01, 0.015),}
\end{align*}
where $\Sigma_i$ represents the uncertainty distribution of the i-th obstacle $\objectconfig^i=\left[x^i, y^i, \phi^i, l_1^i, l_2^i\right]^T$.
The goal of the robot is to find a path leading to the goal region to the left of the workspace while accounting for a given \gls{cp} constraint $p_{\text{max}}\in\{10^{-1},10^{-2},10^{-3}\}$.

The planner assumes a bicycle model for the robot and plans using Hybrid-A$^*$ using a set of 10 motion primitives. 
To check the satisfaction of the \gls{cp} constraint by evaluating if the sate is below the lower bound of the 95\% confidence interval around the mean of 10 \gls{dcpf} Networks. 
We consider two baselines to compare against our method. The first is a common Simple Monte Carlo baseline~\cite{lambert_collision_2008}, also used in~\cite{schmerling_evaluating_2017,banfi_path_2020}. Specifically, we compute the \gls{cp} estimate via Simple Monte Carlo and use a one sided z-test to check if it can be concluded, with $95\%$ confidence, that the \gls{cp} constraint is not violated. The second baseline is the \gls{sprt} approach introduced in~\cite{banfi_path_2020}. In both cases, we run experiments for $p_{\text{max}} \in \{10^{-1},10^{-2},10^{-3}\}$, and examine the impact of using a limited number of samples for testing. 
In the first scenario we use a maximum of $4\cdot 10^6$ samples  for the \gls{sprt} baseline and tested different sample maxima of $\{10^2, 10^3, 10^4, 10^5,10^6,4\cdot 10^6\}$ for the Z-Test.

\begin{figure}[t]
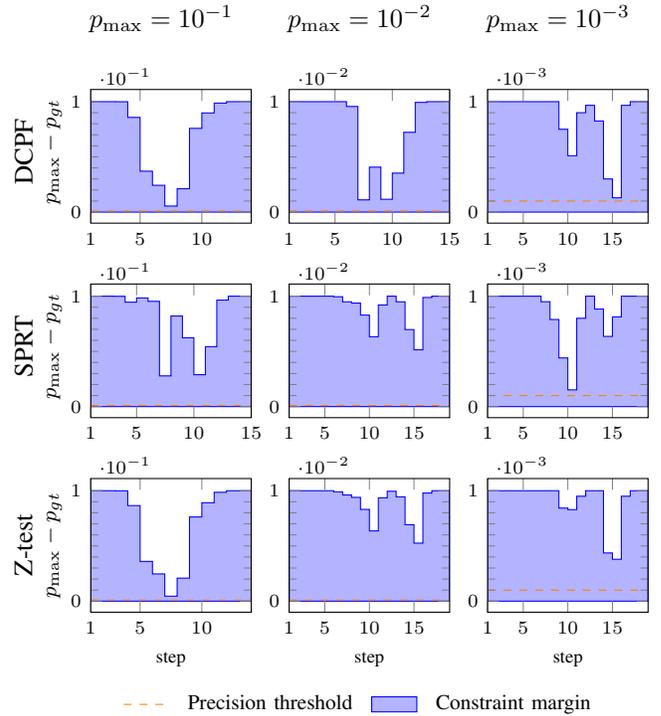

    \centering
    \input{img/experiments/02_obstacles/dcpf/data}
    \input{img/experiments/02_obstacles/sprt/data}
    \input{img/experiments/02_obstacles/ztest/data}
    \setlength{\tabcolsep}{0pt}
    \begin{tabular}{clll}
        &  \multicolumn{1}{c}{$p_{\max} = 10^{-1}$} 
        &  \multicolumn{1}{c}{$p_{\max} = 10^{-2}$} 
        &  \multicolumn{1}{c}{$p_{\max} = 10^{-3}$}  \\ 
        \\
        \raisebox{3\normalbaselineskip}[0pt][0pt]{\rotatebox[origin=c]{90}{\parbox{3cm}{\centering DCPF\\ \footnotesize $p_{\max} - p_{gt}$}}}
        &
        \input{img/experiments/02_obstacles/dcpf/const_1}
        &
        \input{img/experiments/02_obstacles/dcpf/const_2}
        &
        \input{img/experiments/02_obstacles/dcpf/const_3}
        \\
        \raisebox{3\normalbaselineskip}[0pt][0pt]{\rotatebox[origin=c]{90}{\parbox{3cm}{\centering SPRT\\ \footnotesize $p_{\max} - p_{gt}$}}}
        &
        \input{img/experiments/02_obstacles/sprt/const_1}
        &
        \input{img/experiments/02_obstacles/sprt/const_2}
        &
        \input{img/experiments/02_obstacles/sprt/const_3} \\
        \raisebox{3.9\normalbaselineskip}[0pt][0pt]{\rotatebox[origin=c]{90}{\parbox{3cm}{\centering Z-test\\ \footnotesize $p_{\max} - p_{gt}$}}}
        &
        \input{img/experiments/02_obstacles/ztest/10000/const_1}
        &
        \input{img/experiments/02_obstacles/ztest/10000/const_2}
        &
        \input{img/experiments/02_obstacles/ztest/10000/const_3}
        \\
        & \multicolumn{3}{c}{\ref{dcpfconstlegend}} \\
    \end{tabular}
    \caption{The distance of the \gls{cp} constraint $p_{\max}$ to the \gls{cp} $p_{gt}$, computed with the \gls{clt}-based method outlined in \ref{sec:data-gen}, during the planned trajectories seen in figure \ref{fig:a_star_paths}. More conservative methods will have a higher minimal distance. For values below the precision threshold it cannot be verified whether the constraint is respected, due to the uncertainty of $p_{gt}$}
    \label{fig:constraint_margin}
    \vspace{-2em}
\end{figure}

From Figure~\ref{fig:a_star_paths} we can see that the trajectory computed by the planner changes the homotopy class for collision probabilities smaller than $10^{-3}$. Instead, both Z-test and \gls{sprt} change the homotopy class for lower constraints. This happens as both methods will mark a state as unsafe if they cannot decide with the given sample budget if the \gls{cp} constraint is respected. This behavior is highlighted in Figure \ref{fig:constraint_margin}, showing that our method gets closer to the constraint, while not violating it, whereas the other methods stay further away, not allowing the car to pass through the narrow gap.

\begin{figure}[b]
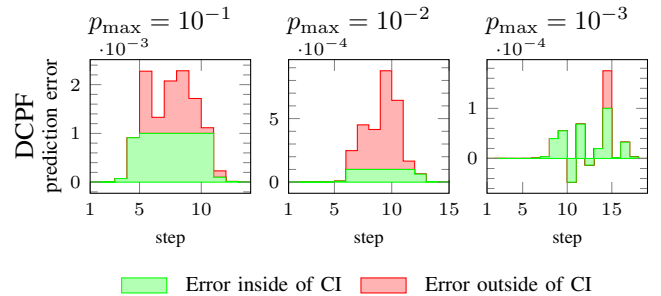

    \vspace{-1em}
    \centering
    \input{img/experiments/02_obstacles/dcpf/data}
    \input{img/experiments/02_obstacles/sprt/data}
    \input{img/experiments/02_obstacles/ztest/data}
    \setlength{\tabcolsep}{0pt}
    \begin{tabular}{cccc}
        & $p_{\max} = 10^{-1}$  &  $p_{\max} = 10^{-2}$  &  $p_{\max} = 10^{-3}$  \\ 
        \raisebox{3.9\normalbaselineskip}[0pt][0pt]{\rotatebox[origin=c]{90}{\parbox{3cm}{\centering DCPF\\ \footnotesize prediction error}}}
        &
        \input{img/experiments/02_obstacles/dcpf/error_1}
        &
        \input{img/experiments/02_obstacles/dcpf/error_2}
        &
        \input{img/experiments/02_obstacles/dcpf/error_3}
        \\
        & \multicolumn{3}{c}{\ref{dcpferrorlegend}} \\
    \end{tabular}
    \caption{Error of \gls{cp} estimate during the planned trajectory for \gls{dcpf} in figure \ref{fig:a_star_paths} The green bar represents an estimation error inside the confidence interval of the ground truth calculation, while the red bar represents an error outside the confidence interval.}
    \label{fig:error_bars}
\end{figure}

\begin{table}[b]
    \vspace{-1em}
    \centering
    \caption{Random obstacles in map experiment results}
    \input{tables/stat_map_obstacles}
    \label{table:statmapobstacles}
\end{table}

We further analyze the behavior of our algorithm by computing the difference $\hat{p}_{\text{coll}} - \bar{p}$, where $\bar{p}$ is the ground truth value, computed using the z-test with an unlimited amount of samples. 
The results, presented in Figure~\ref{fig:error_bars} show that our approach in general computes accurate collision probabilities while finding a less conservative path than Z-test with limited computation. Indeed, the prediction mostly falls into the confidence interval of the ground truth, meaning we cannot distinguish the error from noise in the data generation process. In the experiments, our approach always outputs conservative \gls{cp} estimates thanks to the confidence interval upper bound, reflected in a positive error in the \gls{cp} estimate when the error is outside the ground truth confidence interval. This result is particularly important to provide the required level of safety.

We also test our approach in a realistic scenario and present the results in Table~\ref{table:statmapobstacles}.
In this second experiment, we consider 10 different cleaned-up occupancy maps of real-world environments~\cite{whiting2007topology,aydemir2012can,amigoni2018improving} populated with 120 random obstacles per free \SI{100}{\meter\squared}.
Width and height are between \SI{0.1}{\meter} and \SI{3}{\meter}. The uncertainties are drawn uniformly between 0.001 and 0.1.
The start and goal position are chosen, so that the distance without considering the map is between \SI{35}{\meter} and \SI{40}{\meter}. 
We set the \gls{cp} constraint $p_{\max}$ to $10^{-3}$ and compare our method against s-Test and \gls{sprt} baselines in 10 random configurations per map. The sample maxima are set to $10^6$ for the z-Test and $4\times10^6$ for the \gls{sprt} baseline.
Due to the set computing time limitations the z-Test and \gls{sprt} baselines find fewer solutions. The mean and standard deviation of the path length and planning time are only computed for the paths, where all of the methods found a solution. Also here the \gls{dcpf} approach outperforms the baselines in all metrics.

\begin{figure}[t]
    \centering
    \begin{tabular}{ccc}
    \includegraphics[width=0.9\columnwidth]{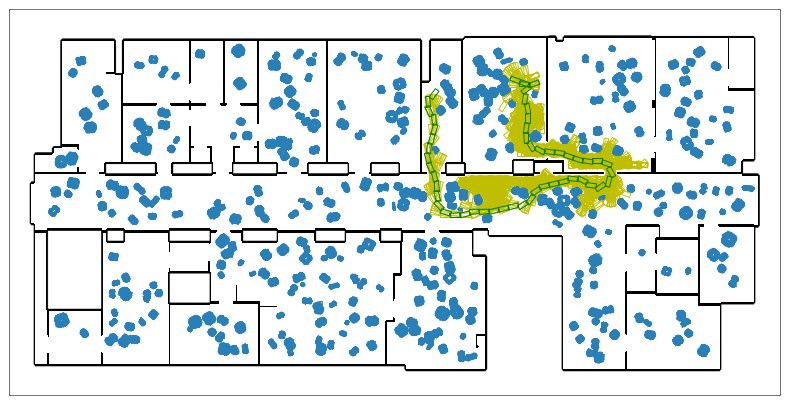} & & 
    \end{tabular}
    \caption{Path found in between random obstacles in an occupancy map with hybrid A* using \gls{dcpf} to ensure satisfaction of the \gls{cp} constraint $p_{\max}$}
    \label{fig:obstacles_map}
\end{figure}

\begin{figure}[t]
    \centering\centering
    \begin{tabular}{c}
        \fbox{\includegraphics[width=0.9\columnwidth]{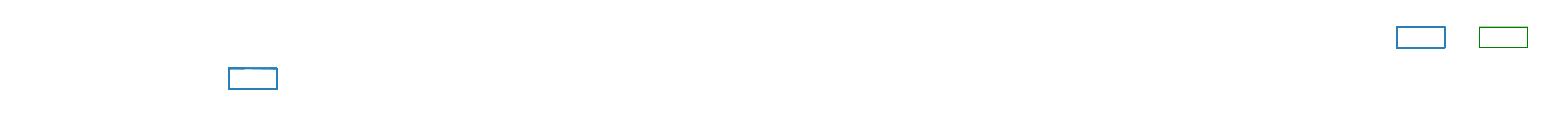}}  \\
        \fbox{\includegraphics[width=0.9\columnwidth]{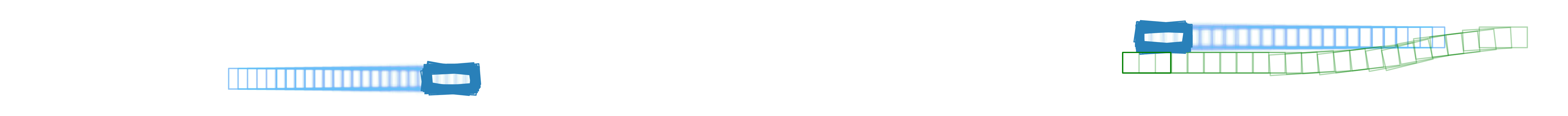}} \\
        \fbox{\includegraphics[width=0.9\columnwidth]{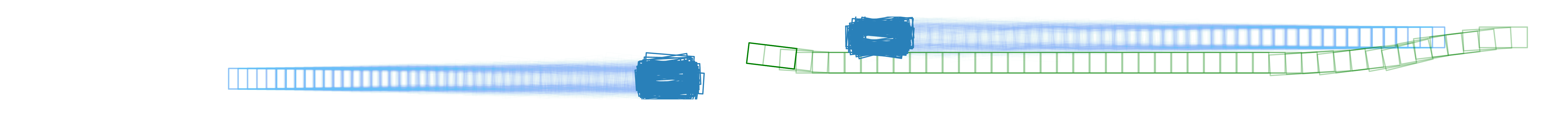}} \\
        \fbox{\includegraphics[width=0.9\columnwidth]{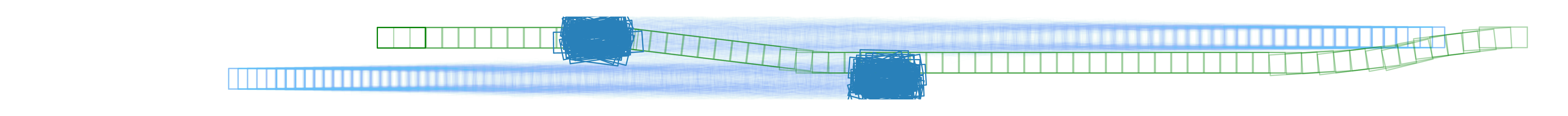}} \\
        \fbox{\includegraphics[width=0.9\columnwidth]{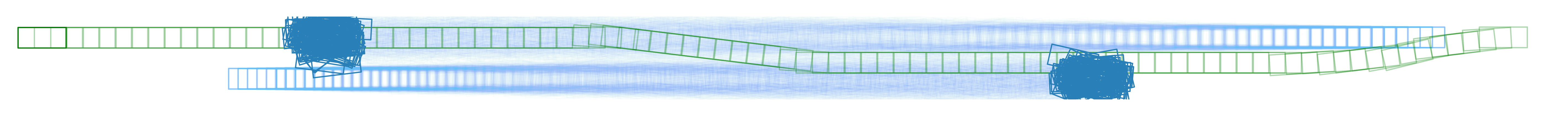}} \\
    \end{tabular}
    \caption{Example of the paths computed during the dynamic overtake experiments}
    \label{fig:overtake}
    \vspace{-2em}
\end{figure}

\paragraph{Dynamic Obstacles} To test the performance with dynamic obstacles we implemented a time-dependent version of Hybrid-A$^*$, and we tested our approach in a dynamic overtake setting, depicted in Fig~\ref{fig:overtake}. The objective of the task is to reach the end of the lane as soon as possible, by overtaking a car running in front of the controlled vehicle, while another car, driving in the opposite lane, is approaching. The position of the non-controlled car is known with uncertainty. The uncertainty grows over time as it would happen if a Kalman filter computes the prediction.

\begin{table}[b]
    \vspace{-1em}
    \centering
    \caption{Overtake experiment results}
    \input{tables/overtake_results}
    \label{table:overtake}
\end{table}

We test this challenging planning task under \gls{cp} constraint $p_{\text{max}}\in\{.1, .01, .001\}$, and we provide the results in Table~\ref{table:overtake}. In this table, "overtake before" and "overtake after" refer to an overtake happening, respectively, before or after the car riding on the opposite lane reaches the one in front of the agent.
The results show a clear effect on the planned overtake for different \gls{cp} constraints. Lowering the budget forces the car to wait for the other car to pass by before the overtake is complete more often, as a result of a bigger safety margin. For the same reasons, the number of episodes where no overtake happens increases when tightening the \gls{cp} constraint. These results show that the proposed task is quite challenging for the \gls{cp} evaluation.
Despite the difficulty of the task, our method provides an accurate estimate of \gls{cp} for every $p_{\text{max}}$ level and, in the experiment setting, we have no constraint violations. This accurate estimation can also be seen by inspecting the values of the mean absolute error and the mean violation, which are also extremely low. 

To improve the relevance of our evaluation, we again use 10 different occupancy maps populated with obstacles moving in rectangular paths or back and forth. For the starting and goal positions, 250 configurations are chosen, such that a significant portion of the map has to be traversed and various moving obstacles have to be avoided. We set the \gls{cp} constraint $p_{\max}$ to $10^{-3}$ and the maximum sample counts for the z-Test and \gls{sprt} baselines to $4\times 10^6$.
Table~\ref{table:dynmapobstacles} shows the results, which show that while the baselines seem to find shorter paths, the planning time is significantly shorter.

\begin{table}[b]
    \vspace{-1em}
    \centering
    \caption{Dynamic obstacles in map experiment results}
    \input{tables/dyn_map_obstacles}
    \label{table:dynmapobstacles}
\end{table}

\begin{figure}[t]
    \centering
    \includegraphics[width=\columnwidth, clip]{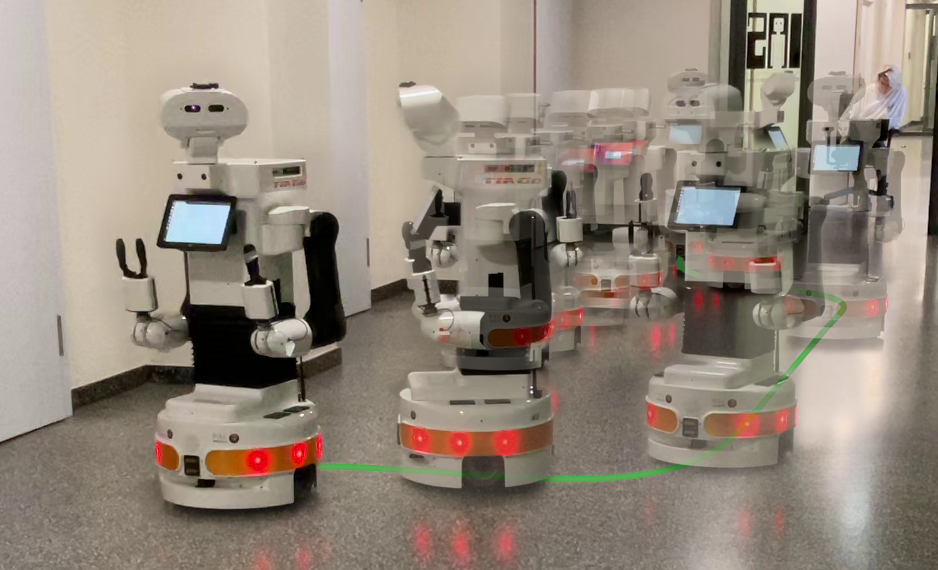}
    \caption{TIAGo robot performing an overtake by planning under uncertainty and collision probability bounds}
    \label{fig:TIAGo}
    \vspace{-2em}
\end{figure}

\paragraph{Real robot experiments} We also provide a small proof-of-concept experiment, where we deploy our approach on a real robot setup. Here we implement the overtake experiment using two TIAGo robots, as shown in Figure~\ref{fig:TIAGo}. The TIAGo robot that is running in front of the controlled one is changing the width of his arms, enlarging his footprint, making his bounding box uncertain in every instant.
To allow for an easy deployment and exploit the advantages of the parallelization coming from our method, we implement a simple motion-primitive-based planner. These motion primitives are pre-generated spline-based overtake trajectories, followed by a simulated bicycle drive controller to estimate the timestamp at each point. We then evaluate them at different discretization levels to quickly narrow down options while ensuring that the chosen trajectory does not violate the \gls{cp} constraint.
The algorithm runs real-time on the system and the overtake happens without any collision happening. In future works, we expect to deploy \gls{dcpf} planners in more complex robotics setups with ease.

%% file: tables/network_ablation_ci_percentage.tex
\begin{tabular}{l|cccc}
\hline
Network &  $[0.0, 10^{-2})$ & $[10^{-2}, 10^{-1})$ & $[10^{-1}, 1]$ & $[0.0, 1.0]$ \\ 
\hline
3x128 & $0.666045$ & $0.602677$ & $0.906839$ & $0.723093$ \\
3x512 & $0.407508$ & $0.569119$ & $0.888136$ & $0.594182$ \\
4x512 & $0.757707$ & $0.701193$ & $0.939518$ & $0.798590$ \\
4x1024 & $0.761130$ & $0.667497$ & $0.934856$ & $0.790180$ \\
5x1024 & $0.770757$ & $0.722915$ & $0.946965$ & $0.812132$ \\
6x1024 & $\mathbf{0.779680}$ & $\mathbf{0.747732}$ & $\mathbf{0.951831}$ & $\mathbf{0.823841}$ \\
7x1024 & $0.757662$ & $0.698077$ & $0.947665$ & $0.800255$ \\
\hline
\end{tabular}

%% file: tables/execution_speeds_reciproc.tex
\footnotesize
\begin{tabular}{l|cccc}
\hline
\multirow{2}{*}{Method} & {Mean Time} & {Std Time} & {Max Time} \\
& [ms] & [ms] & [ms] \\
\hline
SPRT $p_{\max} = 0.1$    & 0.558  & 0.589  & 19.123 \\
SPRT $p_{\max} = 0.01$   & 0.900  & 0.743  & 12.392 \\
SPRT $p_{\max} = 0.001$  & 9.378  & 7.227  & 114.030 \\
z-Test $p_{\max} = 0.1$  & 0.918  & 25.520 & 1761.018 \\
z-Test $p_{\max} = 0.01$ & 3.634  & 49.207 & 2409.020 \\
z-Test $p_{\max} = 0.001$& 68.661 & 598.436& 8554.228 \\
Network CPU $b=1$        & 14.770 & 3.477  & 122.731 \\
Network GPU $b=1$        & 2.636  & 13.067 & 415.570 \\
Network CPU $b=16$       & 1.523  & 0.299  & 10.774 \\
Network GPU $b=16$       & 0.181  & 0.908  & 28.870 \\
Network CPU $b=1024$     & 0.718  & 0.007  & 0.878 \\
Network GPU $b=1024$     & 0.011  & 0.013  & 0.412 \\
\hline
\end{tabular}
\normalsize

%% file: tables/stat_map_obstacles.tex
\footnotesize
\begin{tabular}{p{1.8cm}|rrrr} 
    \hline
    \multirow{2}{*}{Metric} & \multirow{2}{*}{DCPF}    & $z$-Test  & SPRT \\
    & & \tiny  $n=1e6$ & \tiny  $n=4e6$ \\ \hline
    No solution & 23 & 47 & 30 \\
    \multirow{2}{2cm}{Path duration [s]} & 26.133 & 26.133 & 26.101 \\
    & $\pm$6.651& $\pm$6.651 & $\pm$6.680 \\
    \multirow{2}{2cm}{Planning Time[s]} & 335.220 & 6764.291 & 1223.666 \\
    & $\pm$359.047& $\pm$5182.787 & $\pm$869.418 \\
    \hline
\end{tabular}
\normalsize

%% file: tables/overtake_results.tex
\small
\begin{tabular}{p{2.5cm}|rrr}
\hline
$p_{\max}$                                                                 & $10^{-1}$      & $10^{-2}$      & $10^{-3}$   \\ 
\hline
Overtake before & 29 & 29 & 27 \\
Overtake after & 21 & 18 & 18  \\
No overtake & 0 & 3 & 5 \\
Violations & 0 & 0 & 0 \\
\multirow{2}{2cm}{\gls{mae} {\tiny $(\cdot 10^{-3})$}} &
 0.1452 & 0.0187 & 0.0031  \\
 & $\pm$0.0006 & $\pm$0.0001 & $\pm$0.0000 \\ 
 \hline
\end{tabular}
\normalsize

%% file: tables/dyn_map_obstacles.tex
\footnotesize
\begin{tabular}{p{1.8cm}|rrrr} 
    \hline
    \multirow{2}{*}{Metric} & \multirow{2}{*}{DCPF}    & $z$-Test  & SPRT \\
    & & \tiny  $n=4e6$ & \tiny  $n=4e6$ \\ \hline
    No solution & 151 & 192 & 158 \\
    \multirow{2}{2cm}{Path duration [s]} & 173.668 & 146.848 & 159.187 \\
    & $\pm$124.989& $\pm$81.375 & $\pm$102.153 \\
    \multirow{2}{2cm}{Planning Time[s]} & 1342.551 & 6245.175 & 2721.024 \\
    & $\pm$3181.698& $\pm$6432.084 & $\pm$4905.856 \\
    \hline
\end{tabular}
\normalsize

%% file: sections/appendix.tex
\onecolumn
\appendix
\subsection{Max samples computation}
In Section~\ref{sec:data-gen} we set our worst-case maximum total samples to $4\cdot 10^6$. We compute this number as the infimum of the Gaussian test
{\small
\begin{equation*}
    n \geq 1.96^2 \cdot \frac{p\cdot(1-p)}{\epsilon^2}
\end{equation*}}
with
{\small
\begin{align*}
    \epsilon = \begin{cases}
    0.0001 &\quad\text{for}\quad 0 \geq p > 0.01 \\
    0.001  &\quad\text{for}\quad 0.01 \geq p > 0.1 \\
    0.01   &\quad\text{for}\quad 0.1 \geq p \geq 1 \\
    \end{cases}
\end{align*}
}
where $\epsilon$ refers to the accuracy associated with the probability bin. Using these values, the highest lower bound that needs to be fulfilled is at $n \geq 3.803.183$ for the accuracy $\epsilon$ of $0.0001$.
\subsection{Ablation studies}
\begin{figure*}[ht]
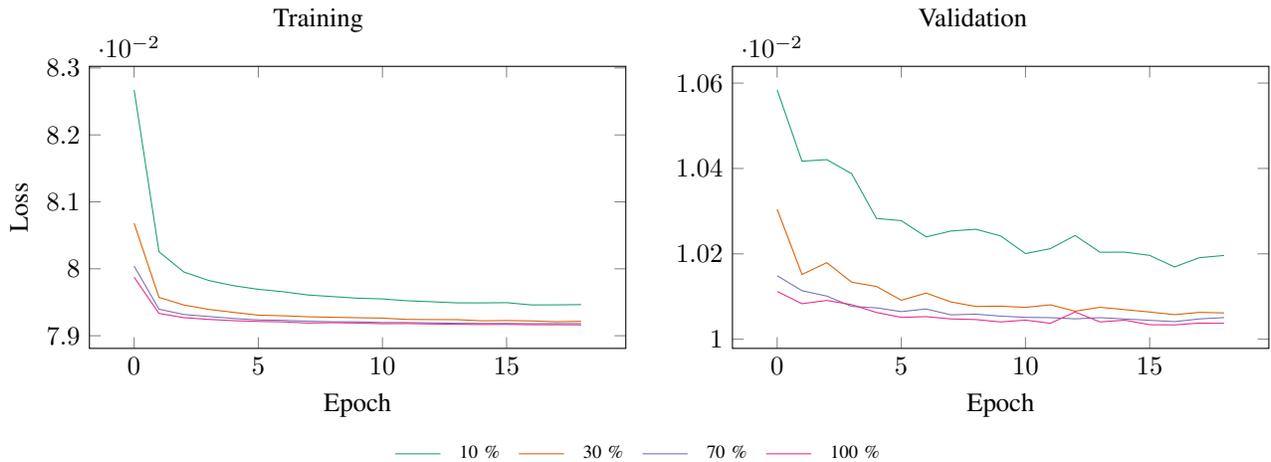

    \centering
    \input{img/experiments/01_network_evaluation/ablations/dataset_size/loss/data}
    \begin{tabular}{cc}
    Training & Validation \\
    \input{img/experiments/01_network_evaluation/ablations/dataset_size/loss/training} & 
    \input{img/experiments/01_network_evaluation/ablations/dataset_size/loss/validation}    
    \end{tabular}
    \ref{datalosslegend}
    \caption{Loss for different dataset sizes}
    \label{fig:data_ablation_loss}
\end{figure*}

\begin{figure}[ht]
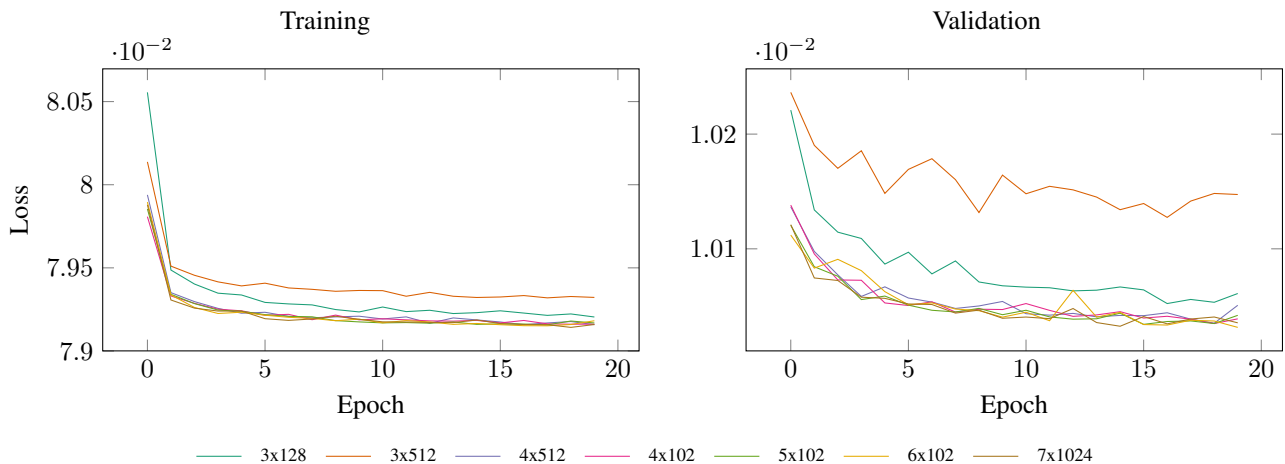

    \centering
    \input{img/experiments/01_network_evaluation/ablations/network_size/loss/data}
    \begin{tabular}{cc}
    Training & Validation \\
    \input{img/experiments/01_network_evaluation/ablations/network_size/loss/training} & 
    \input{img/experiments/01_network_evaluation/ablations/network_size/loss/validation} 
    \end{tabular}
    \ref{networklosslegend}
    \caption{Loss for different network sizes}
    \label{fig:network_ablation_loss}
\end{figure}

\begin{figure*}[ht]
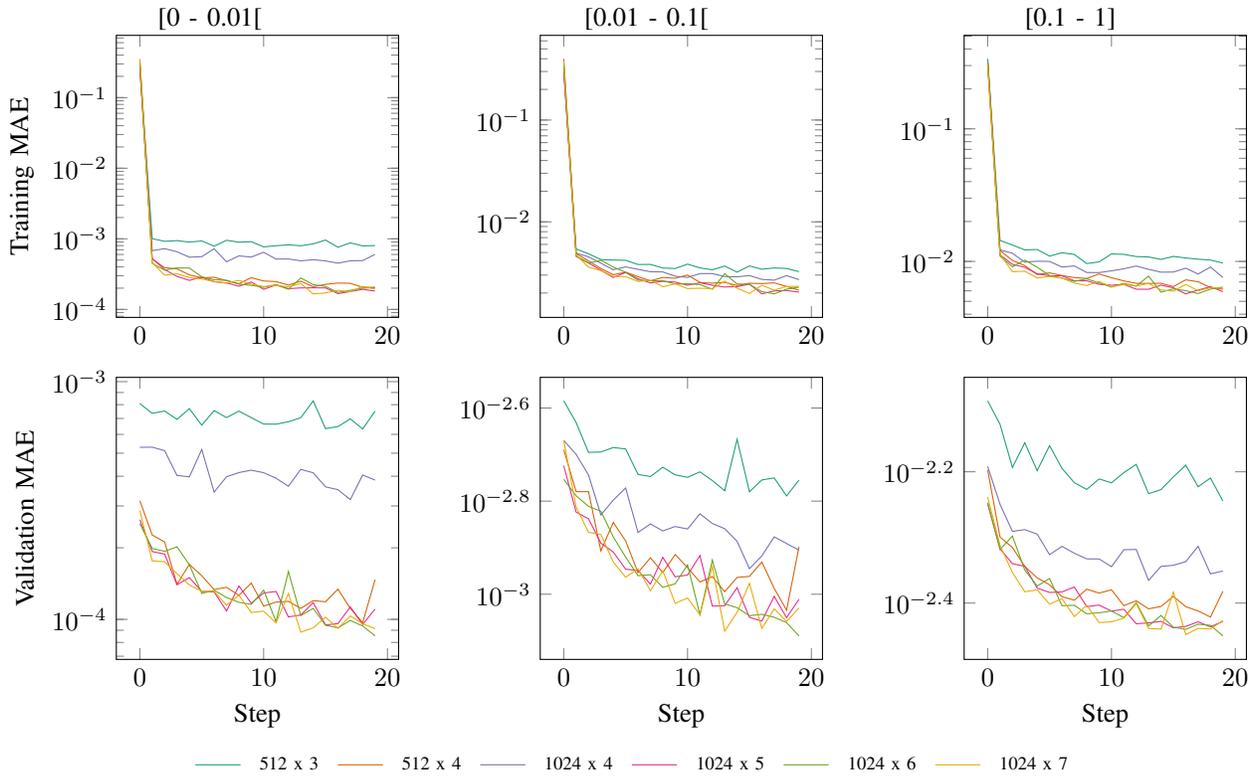

    \input{img/experiments/01_network_evaluation/ablations/network_size/mae/data}
    \centering
    \begin{tabular}{rrr}
    \multicolumn{1}{c}{[0 - 0.01[} & \multicolumn{1}{c}{[0.01 - 0.1[} & \multicolumn{1}{c}{[0.1 - 1]}  \\
    \input{img/experiments/01_network_evaluation/ablations/network_size/mae/training_2}
    &
    \input{img/experiments/01_network_evaluation/ablations/network_size/mae/training_1}
    &
    \input{img/experiments/01_network_evaluation/ablations/network_size/mae/training_0}
     \\
    \input{img/experiments/01_network_evaluation/ablations/network_size/mae/validation_2}
    &
    \input{img/experiments/01_network_evaluation/ablations/network_size/mae/validation_1}
    &
    \input{img/experiments/01_network_evaluation/ablations/network_size/mae/validation_0}
    \end{tabular}
    \ref{networkabslegend}
    \caption{\gls{mae} on train and validation dataset for different network sizes}
    \label{fig:mae_ablation_network}
\end{figure*}

This section presents the results of the ablation studies performed in the network. Specifically, we ablate the dataset size showing the training and validation loss in Fig.~\ref{fig:data_ablation_loss} and the \gls{mae} in Fig.~\ref{fig:data_mae_loss}. We also ablate the network size, showing in Fig.~\ref{fig:network_ablation_loss} the training and validation loss and the respective \gls{mae} in Fig.~\ref{fig:mae_ablation_network}. results show that our method is sufficiently reliable even with smaller dataset sizes and that it is not too dependent on the network size, provided sufficient network capacity.

\begin{figure*}[ht]
    \input{img/experiments/01_network_evaluation/ablations/dataset_size/mae/data}
    \centering
    \begin{tabular}{rrr}
    \multicolumn{1}{c}{[0 - 0.01[} & \multicolumn{1}{c}{[0.01 - 0.1[} & \multicolumn{1}{c}{[0.1 - 1]}  \\
    \input{img/experiments/01_network_evaluation/ablations/dataset_size/mae/training_2}
    &
    \input{img/experiments/01_network_evaluation/ablations/dataset_size/mae/training_1}
    &
    \input{img/experiments/01_network_evaluation/ablations/dataset_size/mae/training_0}
     \\
    \input{img/experiments/01_network_evaluation/ablations/dataset_size/mae/validation_2}
    &
    \input{img/experiments/01_network_evaluation/ablations/dataset_size/mae/validation_1}
    &
    \input{img/experiments/01_network_evaluation/ablations/dataset_size/mae/validation_0}
    \end{tabular}
    \ref{datasetabslegend}
    \caption{\gls{mae} on train and validation dataset for different dataset sizes}
    \label{fig:data_mae_loss}
\end{figure*}

\clearpage

\subsection{Additional details from experiments}
\begin{figure*}[ht]
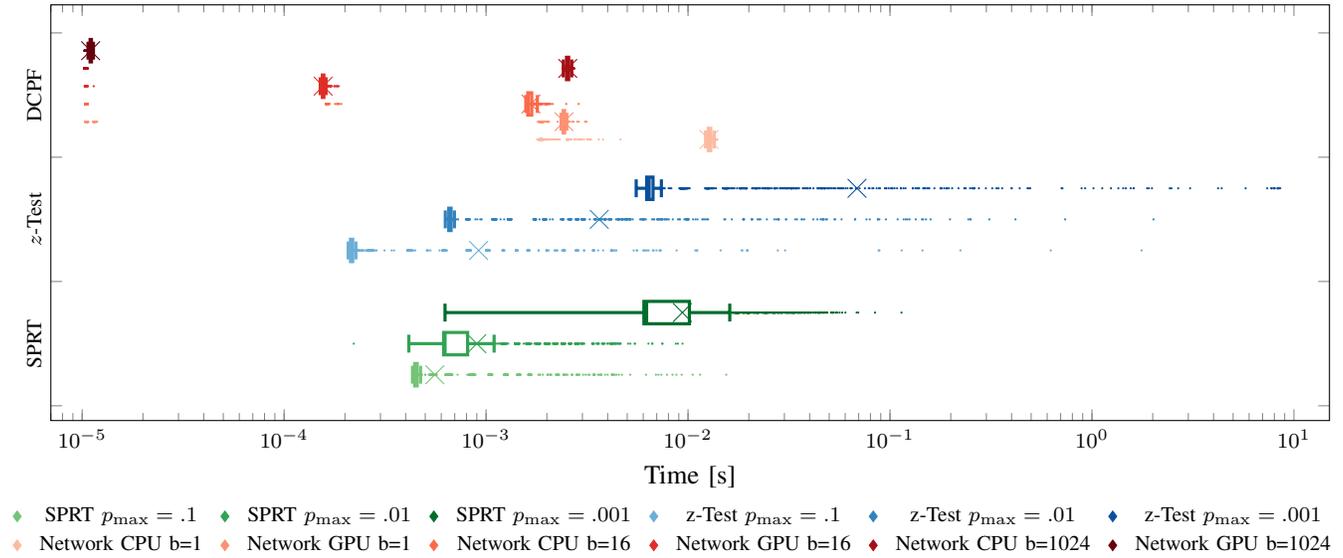

    \centering
    \input{img/experiments/01_network_evaluation/speed_data}
    \input{img/experiments/01_network_evaluation/speed_boxplot}
    \ref{speedlegend}
    \caption{Box plot of inference computation time (per sample), where b is the number of points evaluated in parallel and $p_{max}$ is the probability constraint. If not otherwise specified b=1. The distribution of \gls{cp} is approximately reciprocal. In the plot, outliers are represented by points and the mean of the distribution is represented by an x mark.}
    \label{fig:speed_box}
    \vspace{-0.3cm}
\end{figure*}

In this section, we report additional details from the experimental section presented in the main paper. Figure~\ref{fig:speed_box} presents the runtime experiments as a boxplot. Tables~\ref{table:random_obstacles_01},~\ref{table:random_obstacles_001} and~\ref{table:random_obstacles_0001} instead report more detailed results for the random obstacle experiment for different levels of $p_\text{max}$ with 50 obstacles in an otherwise empty square map of \SI{50}{\meter} by \SI{50}{\meter}. Notice that the computational advantages of \gls{sprt} vanishes for very low \gls{cp} budgets.
\vspace{2cm}

\begin{table}[ht]
    \centering
    \caption{Results from the random obstacle experiments for $p_\text{max}=0.1$}
    \begin{tabular}{lrrr}
    \toprule
     & {DCPF} & \begin{tabular}[c]{@{}r@{}}z-Test\\ $n=1000$\end{tabular} &   \begin{tabular}[c]{@{}r@{}}z-Test\\ $n=10000$\end{tabular}\\
    \midrule
    No solution found       & 24 & 24 & 24 \\
    Mean path length [m]    & 32.6711 & 32.5658 & 32.6579 \\
    Std path length [m]     & 4.2901 & 3.9714 & 4.1724  \\
    Mean time [s]           & 106.2414 & 20.4188 & 131.9221  \\
    \midrule
     &  \begin{tabular}[c]{@{}r@{}}z-Test\\ $n=100000$\end{tabular} & \begin{tabular}[c]{@{}r@{}}z-Test\\ $n=1000000$\end{tabular} & \begin{tabular}[c]{@{}r@{}}SPRT\\ $n=4000000$\end{tabular} \\
    \midrule
    No solution found       & 25 & 25 & 24 \\
    Mean path length [m]    & 32.4400 & 32.4400 & 32.8289 \\
    Std path length [m]     & 3.8201 & 3.8201 & 5.0923 \\
    Mean time [s]           & 1277.1769 & 4890.3260 & 20.0671 \\
    \bottomrule
    \end{tabular}
    \label{table:random_obstacles_01}
\end{table}

\begin{table}[ht]
    \centering
    \caption{Results from the random obstacle experiments for $p_\text{max}=0.01$}
    \begin{tabular}{lrrr}
    \toprule
     & {DCPF} & \begin{tabular}[c]{@{}r@{}}z-Test\\ $n=1000$\end{tabular} &   \begin{tabular}[c]{@{}r@{}}z-Test\\ $n=10000$\end{tabular}\\
    \midrule
    No solution found       & 24 & 24 & 24 \\
    Mean path length [m]    & 33.1316       & 33.0133   & 33.1200 \\
    Std path length [m]     & 4.8050        & 4.9652    & 5.2712 \\
    Mean time [s]           & 121.3385      & 35.7087   & 110.7654 \\
    \midrule
     &  \begin{tabular}[c]{@{}r@{}}z-Test\\ $n=100000$\end{tabular} & \begin{tabular}[c]{@{}r@{}}z-Test\\ $n=1000000$\end{tabular} & \begin{tabular}[c]{@{}r@{}}SPRT\\ $n=4000000$\end{tabular} \\
    \midrule
    No solution found       & 25            & 25        & 19 \\
    Mean path length [m]    & 33.1333 & 33.1333 & 33.9753 \\
    Std path length [m]     & 5.2823 & 5.2823 & 6.0878 \\
    Mean time [s]           & 907.5650 & 3499.4843 & 39.4890 \\
    \bottomrule
    \end{tabular}
    \label{table:random_obstacles_001}
\end{table}

\begin{table}[ht]
    \centering
    \caption{Results from the random obstacle experiments for $p_\text{max}=0.001$}
    \begin{tabular}{lrrr}
    \toprule
     & {DCPF} & \begin{tabular}[c]{@{}r@{}}z-Test\\ $n=1000$\end{tabular} &   \begin{tabular}[c]{@{}r@{}}z-Test\\ $n=10000$\end{tabular}\\
    \midrule
    No solution found       & 21 & 23 & 23 \\
    Mean path length [m]    & 33.7089 & 33.6234 & 33.4286 \\
    Std path length [m]     & 5.3205 & 5.3623 & 5.4877 \\
    Mean time [s]           & 140.6172 & 284.8185 & 327.5759 \\
    \midrule
     &  \begin{tabular}[c]{@{}r@{}}z-Test\\ $n=100000$\end{tabular} & \begin{tabular}[c]{@{}r@{}}z-Test\\ $n=1000000$\end{tabular} & \begin{tabular}[c]{@{}r@{}}SPRT\\ $n=4000000$\end{tabular} \\
    \midrule
    No solution found       & 21 & 21 & 22 \\
    Mean path length [m]    & 33.7342 & 33.7342 & 33.2692 \\
    Std path length [m]     & 5.9058 & 5.9058 & 4.4338 \\
    Mean time [s]           & 1090.1761 & 3434.1721 & 269.8011 \\
    \bottomrule
    \end{tabular}
    \label{table:random_obstacles_0001}
\end{table}